\documentclass[12pt,onecolumn,journal]{IEEEtran}
\usepackage{booktabs}
\usepackage{epsfig}
\usepackage{epstopdf}
\usepackage{graphics}
\usepackage{graphicx}
\usepackage{subfigure}
\usepackage{booktabs}
\usepackage[usenames,dvipsnames]{xcolor}
\usepackage{psfrag}
\usepackage{verbatim}
\usepackage{algorithm}
\usepackage{algorithmic}
\usepackage{amsmath}
\usepackage{amssymb}
\usepackage{amsthm}
\usepackage{bm}
\usepackage{booktabs}
\usepackage{epsfig}
\usepackage{epstopdf}
\usepackage{graphics}
\usepackage{graphicx}
\usepackage{subfigure}
\usepackage{booktabs}
\usepackage[usenames,dvipsnames]{xcolor}
\usepackage{psfrag}
\usepackage{url}
\usepackage{chngcntr}
\usepackage{cite}

\usepackage{multirow}
\usepackage{chngcntr}
\usepackage{cite}

\usepackage{multirow}

\usepackage{colortbl}

\usepackage{fixltx2e}
\usepackage{afterpage}

\newcommand{\parcom}[1]{}

\definecolor{r}{cmyk}{0,0.87,0.68,0.32}
\definecolor{g}{cmyk}{0.87,0,0.68,0.32}
\definecolor{s}{cmyk}{0.435,0.435,0.68,0.32}
\counterwithin{paragraph}{section}

\def\y{\mathbf{y}}
\def\f{\mathbf{f}}
\def\z{\mathbf{z}}
\def\x{\mathbf{x}}

\def\L{\mathcal{L}}

\def\S{\mathcal{S}}

\def\V{\mathcal{V}}
\def\What{\hat{\mathbf{W}}}
\def\W{{\mathbf{W}}}
\def\yhat{\hat{y}}

\def\F{\mathbf{F}}
\def\F{\mathbf{F}}

\newcommand{\cred}[1]{{#1}}

\author{Filipe Condessa, \emph{Student Member, IEEE}, Jos\'{e} Bioucas-Dias,
  \emph{Senior Member, IEEE}, Carlos A. Castro, John A. Ozolek, and Jelena Kova\v{c}evi\'{c}, \emph{Fellow, IEEE}
    \thanks{Filipe Condessa is with Instituto de
      Telecomunica\c{c}\~oes and the Dept. of Electrical and Computer
      Engineering at Instituto Superior T\'ecnico, Universidade de
      Lisboa, Portugal, and the
      Dept. of Electrical and Computer Engineering at Carnegie Mellon
      University, Pittsburgh, PA, condessa@cmu.edu.  Jos\'{e}
      Bioucas-Dias is with Instituto de Telecomunica\c{c}\~oes and
      Dept. of Electrical and Computer Engineering at Instituto
      Superior T\'ecnico, Universidade de Lisboa, Portugal,
      bioucas@lx.it.pt.  Carlos Castro is with the Dept. of Obstetrics
      and Gynecology, Magee-Womens Research Inst., Foundation Univ. of
      Pittsburgh, Pittsburgh, PA, ccastro@mwri.magee.edu.  John Ozolek
      is with the Dept. of Pathology, Children's Hospital of
      Pittsburgh, University of Pittsburgh School of Medicine,
      Pittsburgh, PA, ozolja@upmc.edu.  Jelena Kova\v{c}evi\'{c} is
      with the Dept. of Electrical and Computer Engineering, and the Dept. of Biomedical
      Engineering at Carnegie Mellon University, Pittsburgh, PA,
      jelenak@cmu.edu.}  }

\begin{document}

\title{Image Classification with Rejection using Contextual Information}

\markboth{}
{}

\maketitle

\begin{abstract}
 We introduce a new supervised algorithm for image classification with rejection using multiscale contextual information.
 Rejection is desired in image-classification applications that require a robust classifier but not \cred{the classification of} the entire image.
 The proposed algorithm combines local and multiscale contextual information with rejection, improving the classification  performance.

As a probabilistic model for classification, we adopt a multinomial logistic regression.
The concept of rejection with contextual information is implemented by modeling the classification problem as an energy minimization problem over a graph representing local and
  multiscale similarities of the image.
The rejection is introduced through an energy data term associated with the classification risk and the contextual information through an energy   smoothness term associated with the local and multiscale
  similarities within the image.
We \cred{illustrate} the proposed method on \cred{the classification of } images of H\&E-stained teratoma tissues.
\end{abstract}


\IEEEpeerreviewmaketitle




\section{Introduction}
\label{sec:introduction}

\cred{
\parcom{High cost of good training sets}
In many classification problems, the cost of creating a training set that is statistically representative of the input dataset is often high.
This is due to the required size of the training set, and the difficulty of obtaining a correct labeling resulting from unclear   class separability and the possibility of presence of unknown classes.
In this work, we were motivated by the need for automated tissue identification (classification) in images from Hematoxylin and Eosin (H\&E)  stained histopathological slides~\cite{BhagavatulaFKGOCK:10,McCannBYFOK:12,CondessaBCOK:13,BhagavatulaMFCOK:13}.
H\&E staining is used both for diagnosis as well as to gain a better understanding of the diseases and their processes, consisting of the sequential staining of a tissue with two different stains that have different affinities to different tissue components.
}


\parcom{Subclass of image classification problems we attempt to solve}
In this paper, we are interested in a subclass of image classification problems with the following characteristics:
\begin{itemize}
\item The classification is not directly based on the observation of pixel values but on higher-level features;
\item The characteristics of the image make it impossible to have access to pixelwise ground truth, leading to small, unbalanced, noisy, or incomplete training sets;
\item The pixels may belong to unknown classes;
\item The classification accuracy at pixels belonging to interesting or known classes is more important than the classification accuracy  at pixels belonging to uninteresting or unknown classes;
\item The need for high accuracy surpasses the need to classify all
  the samples.
\end{itemize}
 
\subsection{Goal}
\parcom{Rejection yields performance improvements; so does
  exploring spatial and multilevel similarities}
In problems as above, introducing a rejection yields improvements
in the classification performance --- \emph{classification with
  rejection}.  Further improvements in accuracy can be obtained by
exploiting spatial and multilevel similarities ---
\emph{classification using contextual information}.  Our goal is to
\textbf{combine classification with rejection and classification using
  contextual information in an image classification framework} to
obtain improved classification performance.

\subsection{Classification with Rejection}
\parcom{Conceptualization of a classifier with rejection}
A classifier with rejection can be seen as a coupling of two
classifiers: (1) a general classifier that classifies a sample and (2)
a binary classifier that, based on the information available as input
and output of the first classifier, decides whether the classification
performed by the first classifier was correct or incorrect.  As a
result, we are able to classify according to the general classifier,
or reject if the decision of the binary classifier is that the former
classification is incorrect.

\parcom{Theoretical works on classification with rejection}
A classifier with rejection allows for coping with unknown information
and reducing the effect of nonideal training sets.  It was first
analyzed in \cite{Chow:70}, where Chow's rule for optimum error-reject
trade-off was presented.  Based on the posterior probabilities of the
classes given the features for the classification, Chow's rule allows
for the determination of a threshold for rejection, such that the
classification risk is minimized.  The authors in \cite{FumeraRG:00}
point out that Chow's rule only provides the optimal error-reject
threshold if these posterior probabilities are exactly known.  They
propose the combination of class-related reject thresholds to improve
the error-reject trade-off.  Parameters are selected using the
constrained maximization of the accuracy subject to upper bounds on
the rejection rate as a performance metric.  In \cite{HerbeiW:06}, the
authors present a mathematical framework for binary classification
with rejection. In that approach, the rejection is based on risk
minimization and the cost for each different binary classification
error considered.


\parcom{Rejection from combination of multiple classifier outputs}
Usually, the rejection is applied as a \emph{plug-in} rule to the
outputs of a classifier. It is also possible, however, to combine the
output of multiple classifiers (multiple general classifiers) to
create rejection.  In \cite{FogiaSTV:99}, the authors present a
multi-expert system based on a Bayesian combination rule. The
reliability of the classification is estimated from the posterior
probabilities of the two most probable classes, and the rejection
works by thresholding the reliabilities.


\parcom{Embedding rejection in the classifier}
Another approach is to include the rejection in the classifier itself
as an \emph{embedded rejection} instead of a \emph{plug-in} rule.  In
\cite{FumeraR:02}, the rejection is embedded in a Support Vector
Machine (SVM), in which the rejection is present in the training phase
of the SVM and included in the formulation in close association with
the separating hyperplane resulting from the SVM.  This leads to a
nonconvex optimization problem that can be approximately solved by
finding a surrogate loss function.  In \cite{BartlettW:08} and
\cite{YuanW:10}, the statistical properties of a surrogate loss
function are studied and applied to the task of rejection by risk
minimization. In \cite{Wegkamp:07}, the use of LASSO-type penalty for
risk minimization is analyzed.

\parcom{Rejection as a second classifier}
Yet another approach consists in having a second classifier with
access to the input and output of the first classifier instead of a
\emph{plug-in} rule or an \emph{embedded rejection}.  In
\cite{FogiaPSV:07}, the second classifier is trained with the main
classifier to assess the reliability of the main classifier. The
rejection is based on thresholding the reliability provided by the
second classifier.

\parcom{Rejection in multilabel classification problems}
More recently, in \cite{PillaiFR:13}, the authors present a framework
for the multilabel classification problem with rejection.  A trade-off
between the accuracy of the nonrejected samples and the rejection cost
is found as a result of a constrained optimization
problem. Furthermore, an application-specific reliability measure of
the classification with rejection inspired on the F-score (weighted
harmonic mean of precision and recall) is defined.

\parcom{What do present wrt. rejection}
In the present work, we propose a classification system with rejection
using contextual information.  To assess the performance of the
method, \cred{in addition to the fraction of rejected samples $r$ and the classification accuracy on the subset of nonrejected samples $A$}, we use the concept of classification quality $Q$ and rejection
quality $\phi$~\cite{CondessaBK:15_rm}.  The classification quality can
be defined as the accuracy of a binary classifier that aims to
classify correctly classified samples as nonrejected and incorrectly
classified samples as rejected.  
Maximizing the classification quality
leads both to keeping correctly classified samples and rejecting
incorrectly classified samples.  The classification quality allows us
to compare different classifiers with different rejection ratios and
accuracies.  The rejection quality can be defined as the positive
likelihood ratio of a binary classifier that aims to classify
correctly classified samples as nonrejected and incorrectly classified
samples as rejected.  It compares the proportion of correctly
classified to incorrectly classified samples in the set of rejected
samples to the proportion on the entire data.  The rejection quality
provides insight into the ability of a classifier with rejection to
concentrate incorrectly classified samples in the set of rejected
samples.


\subsection{Classification with Contextual Information}
\parcom{Basic assumption for classification with contextual information}
The basic assumption for classification with contextual information is
that the data is not spatially independent: in most real-world data,
two neighboring pixels are likely to belong to the same class.  This
assumption can be extended to include multiple definitions of a
neighborhood: local, nonlocal, and multiscale.

\parcom{Contextual information is prevalent in image processing tasks}
The use of contextual information is prevalent in tasks in which the
spatial dependencies play an important role, such as image
segmentation and image reconstruction \cite{BoykovVZ:01}.  In
\cite{KumarH:06}, the authors formulate a discriminative framework for
image classification taking in account spatial dependencies.  This
framework allows both the use of discriminative probabilistic models
and adaptive spatial dependencies.

\parcom{Starting from Hyperspectral. How we use the context tools}
For the purposes of our application, we can learn from hyperspectral
image classification, where the use of of contextual information is
prevalent~\cite{LiBDP:11,BioucasPCSNC:13}.  We model classification
with contextual information as a Discriminative Random Field
(DRF) \cite{KumarH:06} with the \emph{association potential} linked
with the pixelwise class posterior probabilities and the
\emph{interaction potential} linked with a multilevel logistic (MLL)
Markov random field (MRF)~\cite{LiMRF:01} endowed with a neighboring
system associated with a multi-scale similarity graph.  This MLL-MRF
promotes segmentations in which neighboring samples are likely to
belong to the same class at multiple scales, leading to multi-scale
spatial consistency among the classifications.


\subsection{Classification with Rejection Using Contextual Information}
\begin{figure}[ht!]
  \begin{tabular}{c}
      \includegraphics[width=1\columnwidth]{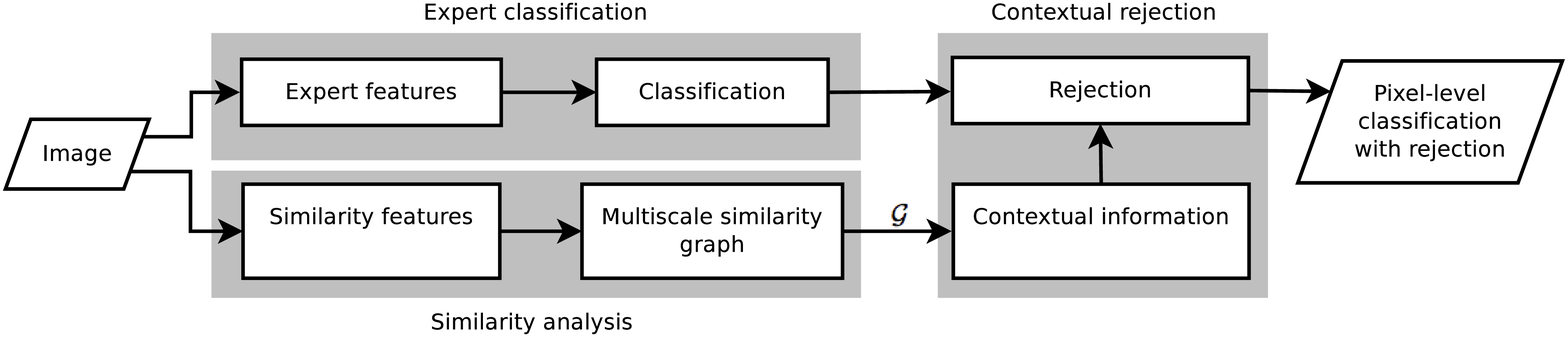}
  \end{tabular}
  \caption{ \label{diag:rejclass} Classification with rejection using
    contextual information. Each gray block is discussed in a separate
    section: similarity analysis in Section~III,
    expert classification in Section~IV, and contextual rejection in
    Section~V. }
\end{figure}

\parcom{Description of proposed framework - top-level}
The proposed framework, shown in Fig.~\ref{diag:rejclass}, combines
classification with rejection with classification with contextual
information.  Our approach allows for not only rejecting a sample when
the information is insufficient to classify, but also for not
rejecting a sample when an "educated guess" is possible based on
neighboring labels (local and nonlocal from the spatial point of
view).  We do so by transforming the soft classification (posterior
distributions) obtained by an expert classifier into a hard
classification (labels) that considers both rejection and contextual
information.

\parcom{Description of proposed framework - middle-level}
An expert classifier is designed based on application-specific
features and a similarity graph is constructed representing the
underlying multiscale structure of the data.  The classification risk
from the expert classifier is computed and the rejection is
introduced as a simple classification risk threshold rule in an
extended risk formulation.  This formulation consists in a maximum a
posteriori (MAP) inference problem defined on the similarity graph,
thus combining rejection and contextual information.

\parcom{Comparison with rejection only}
Compared with classification with rejection only, our approach has an
extra degree of complexity: the rejection depends not only on a
rejection threshold for the classification but also on a rejection
consistency parameter.  By imposing a higher rejection consistency,
the rejected samples become rejection areas (that is, a nonrejected
sample surrounded by rejected samples will tend to be rejected too),
which is meaningful in the task of image classification.

\parcom{Compared with context only}
Compared with classification with contextual information only, this
problem is of the same complexity, as the rejection can be treated as
a class, and class-specific transitions can be easily modeled.




\subsection{Outline of the Paper}
In Section~\ref{sec:background}, we describe the background for our
framework: partitioning, feature extraction, and classification.  In
Section~\ref{sec:similarity}, we explore the similarity analysis block
of the framework and the design of a multilevel similarity graph that
represents the underlying structure of the data.  In Section
\ref{sec:classification}, we describe the \cred{elements of the} expert classification block
of the framework \cred{not described on the background}.  We introduce the
rejection as a mechanism for handling the inability of the classifier
to correctly classify all the samples.  In
Section~\ref{sec:contextual_rejection}, we combine the expert
classification and the multiscale similarity graph in an energy
minimization formulation to obtain classification with rejection
using contextual information.  In Section~\ref{sec:experimental}, we
apply our framework to \cred{classification of} real data: \cred{natural images}, and  H\&E-stained teratoma tissue \cred{images}.
Finally, Section~\ref{sec:conclusion} concludes the paper.

\section{Background}
\label{sec:background}
We now describe the background for our work in terms of image
partitioning, features, classification, and methods used to compute
the MAP solution. 

\parcom{Notation}
Let $\mathcal{S} = \{1, \hdots, s \}$ denote the set of pixel
locations, $\z_i\in\mathbb{R}^d$ denote an observed vector at pixel
$i\in {\cal S}$, $I=[\z_1, \z_2, \dots, \z_s]\in\mathbb{R}^{d\times s}$
denote an observed image, $P = \{\x_1, \hdots, \x_n \}$ denote a
partition of $\cal S$, $\mathcal{V} = \{1, \hdots, n \}$ denote a set
indexing the elements of the partition $P$ termed \emph{superpixels},
and $\mathcal{E} = \mathcal{V}\times \mathcal{V}$ denote a set
indexing pairs of neighboring superpixels. Given that $P$ is a
partition of $\cal S$, then $\x_i \subset {\cal S}$, for $i\in\cal N$,
$ \x_i \cap \x_j = \emptyset$ for $i\neq j\in\cal N$, and $\cup_{i=1}^n
\x_i = \mathcal{S}$.

\subsection{Partitioning}
\parcom{Partitioning: why and how}
To decrease the dimensionality of the problem, and thus the
computational burden, we partition the set of pixel locations $\cal S$
into a partition $P$, allowing for the efficient use of graph-based
methods. The partitioning of the image is performed by
oversegmentation creating superpixels as described in
. This method, as is typical in most
segmentation techniques, aims at maintaining a high level of
similarity inside each superpixel and high dissimilarity between
different superpixels.

\parcom{Characteristics of  partitioning and drawbacks}
Because of how the superpixels are created (measuring the evidence of a boundary between two regions), there is a high degree of inner similarity in each partition element; the elements of a superpixel will very likely belong to the same class.
The major drawback of using this partitioning method is that the partition elements are highly nonuniform in terms of size and shape.

\subsection{Features}
\label{subsec:features}
\parcom{Encoding knowledge in two levels of features. Top level
  view on feature extraction from partitions}
We use two kinds of features: (1) application-specific features encode
expert knowledge and are used to classify each partition element, and
(2) generic similarity features represent low-level similarities of
the image and are used to assess the similarity among the partition
elements.  From each partition element $\x_i$, we extract statistics of the application-specific features and of the similarity
features (from all pixels belonging to the same partition element),
mapping from features defined on an image pixel space to features
defined on an image partition space.

\subsection{Classification}
\label{subsec:classification}
\parcom{Description of classification problem}
Given the partition $ P$ and the associated feature matrix $\F = [\f_1, \hdots, \f_n]$, \cred{with $\f_i \in \mathbb{R}^m$
 the $m$-dimentional application-specific features } ,
 we wish to classify each partition
element $\x_i\in P$ into a single class.  We do so by assigning to it a
label $y_i \in \mathcal{L} = \{1, \hdots , N\}$ representative
of its class.  {This assignment is performed by maximizing the
posterior distribution $p(\y|\F)$ with respect to $\y=[y_1,\dots,y_n]$,
that is, by computing MAP labeling
\begin{equation}
\label{eq:map_pre}
  \hat{\y} \in  \arg\max_{\y\in {\cal L}^n} p(\y | \F) .
\end{equation}
\cred{We note that under the assumption of conditional independence of
  features given the labels $p(\y | \F) = \prod_{i\in \S} p(y_i|\f_i)$ and of equiprobable class probabilities $p(y_i) = p(y_j)$, for all $i,j \in \S$, we can reformulate the MAP formulation in \eqref{eq:map_pre} as
\begin{equation}
\label{eq:map_ext}
  \hat{\y} \in  \arg\min_{\y\in {\cal L}^n} \sum_{i \in \S} -\log p(y_i | \f_i ) - \log p(\y) .
\end{equation}}


\parcom{DRF model adopted for the posterior}
For the posterior $p(\y|\F)$ we adopt the DRF model~\cred{\cite{KumarH:06}},
\begin{equation}
  \begin{split}
    \label{eq:model_DRF}
    p(\y|\F) \propto \exp\bigg( -(1-\alpha)\sum_{i\in\cal V}
    D(y_i,\f_{i}) \\- \alpha\sum_{\{i,j\}\in \cal E}
    V_{\{i,j\}}(y_i,y_j)\bigg),
\end{split}
\end{equation}
where $-D(y_i,\f_{i})$ is the \emph{association potential}, which links
discriminatively the label $y_i$ with the feature vector $\f_{i}$,
$-V_{\{i,j\}}(y_i,y_j)$ is the \emph{interaction potential}, which
models the spatial contextual information, and $\alpha \in [0,1]$ is a
regularization parameter that controls the relative weight of the two
potentials.  The posterior (\ref{eq:model_DRF}) is a particular case
of the DRF class introduced in \cite{KumarH:06}, because the
association potential does not depend on the partition
elements. The DRF model used constitutes an excellent trade-off between model
complexity and goodness of the inferences, as shown in
Section~\ref{sec:experimental}.


\parcom{Association potential of DRF as a MLR}
To completely define (\ref{eq:model_DRF}), we need to specify the association potential $-D$ and the interaction potential $-V_{\{i,j\}}$. 
In this work, we \cred{start from the assumption }  that $-
D(y_i,\f_{i}) = \log p(y_i|\f_{i}, \W)$, \cred{resulting from \eqref{eq:map_ext} and \eqref{eq:model_DRF}}, where $p(y_i|\f_{i}, \W)$ is the \emph{multinomial  logistic regression (MLR)}~\cite{Bohning:92} parameterized with the matrix of regression coefficients $\W$, $-V_{\{i,j\}}(y_i,y_j) = w_{ij}\delta_{y_i,y_j}$, where $w_{ij}\geq 0$ is a weight to be defined late,r and $\delta_{i,j}$ is the Kronecker symbol ({\em i.e.}, $\delta_{i,j} = 1$ if $i=j$ and $\delta_{i,j} = 0 $ if $i\neq j$).
This class of association potentials, which define a MLL-MRF prior \cite{LiMRF:01}, promotes neighboring labels of the same class. 
In the following subsection we address the learning of the MLR
regression matrix $\W$ detail.

\subsubsection{Multinomial Logistic Regression}
\parcom{MLR definition}
Let $k(\f) = [k_0(\f), \hdots, k_q(\f)]^T$ denote a vector of nonlinear
functions $k_i:\mathbb{R}^{m}\rightarrow \mathbb{R}$, for
$i=0,\dots,q$, with $q$ the number of training samples and with $k_0 =
1$.
The MLR models the \emph{a posteriori} probability of
$y_i\in\cal L$ given $\f\in\mathbb{R}^{m}$ as
\begin{equation}
  \label{eq:MLR_prob}
  p(y_i = l | \f,\W) = \frac{e^{\mathbf{w}_l^T k(\f)}}{\sum_{j=1}^N e^{\mathbf{w}_j^T k(\f)}},
\end{equation}
where $\W=[\mathbf{w}_1,\dots,\mathbf{w}_N]\in\mathbb{R}^{(q+1)\times N}$ the matrix of
regression coefficients. Given that $p(y_i | \f,\W)$ is invariant with
respect to a common translation of the columns of $\W$, we arbitrarily
set $\mathbf{w}_N = 0$.


\subsubsection{Learning the Regression Coefficients $W$}
\label{sec:smlr}
\parcom{Description of the inference problem}
Our approach is supervised; we can thus split the dataset into a
training set ${\cal D} =\{ (y_i,\f_{i}),\,i\in{\cal T}\}$, where ${\cal
  T} \subset \cal V$ is a set indexing the labeled superpixels, and the set
$\{\f_{i},\,i\in{\cal V -T}\}$ containing the remaining unlabeled
feature vectors. Based on these two sets and on the DRF model
(\ref{eq:model_DRF}), we can infer matrix $\W$ jointly with the MAP
labeling $\hat{\y}$.  Because it is difficult to compute the
normalizing constant of $p(\y|\F)$, this procedure is complex and
computationally expensive.

\parcom{Sparse model to learn the matrix of regression coefficients}
Aiming at a lighter procedure to learn the matrix $\W$, we adopt the
\emph{sparse multinomial logistic regression} (SMLR) criterion
introduced in \cite{KrishnapuramCFH:05}, which, fundamentally,
consists in setting $\alpha = 0$ in (\ref{eq:model_DRF}), that is,
disconnecting the interaction potential, and computing the MAP
estimate of $\W$ based on the training set ${\cal D}$ and on a
Laplacian independent and identically distributed prior for the
components of $\W$.  We are then led to the optimization
\begin{equation}
  \label{eq:LORSAL_pre}
  \widehat{\W} \in  \arg\max_{W}\,\,l(\W) + \log p(\W),
\end{equation}
with $l(\W) = \sum_{i\in\cal T} \log p(y_i | \f_{i},\W)$ \cred{the log-likelihood}, and $p(\W)
\propto e^{-\lambda \|\W\|_{1,1}}$ \cred{the prior}, where $\lambda$ is the
regularization parameter and $\| \W \|_{1,1}$ denotes the sum of the
$\ell_1$ norm of the columns of the matrix $\W$. 
\cred{The prior $p(\W)$ promotes sparsity on the components of $\W$.}
It is well known that the Laplacian prior (the $\ell_1$ regularizer in the regularization
framework) promotes sparse matrices $\W$, that is, matrices $\W$ with most
elements set to zero.  The sparsity of $\W$ avoids overfitting and thus
improves the generalization capability of the MLR, mainly when the
size of the training set is small \cite{KrishnapuramCFH:05}.  The
sparsity level is controlled by \cred{the parameter} $\lambda$.

\subsubsection{LORSAL}
\parcom{LORSAL is used to solve the optimization associated with
  learning the regression coeffs.}
We use the \emph{logistic regression via variable splitting and
  augmented Lagrangian} (LORSAL) algorithm (see~\cite{LiBDP:11}) to
solve the optimization (\ref{eq:LORSAL_pre}). The algorithm is quite
effective from the computational point of view, mainly when the
dimension of $k\in\mathbb{R}^{q+1}$ is large.

\parcom{LORSAL solving the optimization problem}
LORSAL solves the equivalent problem
\begin{equation}
  \label{eq:LORSAL}
    \min_{\W,\mathbf{\Omega}} -l(\W) + \lambda ||\mathbf{\Omega}||_{1,1}, \quad \textrm{subject to:} \quad \W=\mathbf{\Omega}
\end{equation}
The formulation in \eqref{eq:LORSAL} differs from the one in
\eqref{eq:LORSAL_pre} in the sense that $\log p(\W)$ is replaced by
$\log p(\mathbf{\Omega})$ with the constraint $\W=\mathbf{\Omega}$ added to the
optimization problem, introducing a variable splitting.  Note that
$-l(\W)$ is convex but nonquadratic, and $\lambda ||\mathbf{\Omega}||_{1,1}$ is
convex but nonsmooth, thus yielding a convex nonsmooth and
nonquadratic optimization.  LORSAL approximates $l(\W)$ by a quadratic
upper bound~\cite{Bohning:92}, transforming the nonsmooth convex
minimization \eqref{eq:LORSAL} into a sequence of $\ell_2$-$\ell_1$
minimization problems solved with the \emph{alternating direction
  method of multipliers} \cite{EcksteinADMM:92}.

\cred{
\parcom{RBF as the nonlinear regression function. Instantiation of
  classification choices}
Given a set of indices corresponding to the training samples
$\mathcal{T}$ and its respective training set
$\mathcal{F}_{\mathcal{T}} = \{\F_{j}\}_{\{j \in \mathcal{T}\}}$, a
radial basis function (RBF) is a possible choice of function in the vector of
nonlinear regression function $k$ used in \eqref{eq:MLR_prob}, which
allows us to obtain a training kernel (computed by a RBF kernel
of the training data). 
\cred{This allows us to deal with features that are not linearly separable.} 
{To normalize the values of the nonlinear regression function,} the bandwidth of the RBF
kernel is set to be the square root of the average of the distance
matrix between the training and test sets. 
With both the regressor matrix $\W$ and the nonlinear regression function
$k$ defined, we obtain the class probabilities from the MLR
formulation in \eqref{eq:MLR_prob}.
}

\subsection{Computing the MAP Labeling}
\parcom{Computation of the MAP is an integer optimization problem}
From \eqref{eq:model_DRF}, we can write the MAP labeling optimization
as
\begin{equation}
  \label{eq:class_np_1}
  \arg\min_{\y\in {\cal L}^n} \,\, (1-\alpha)\sum_{i \in \cal V} D(y_i) + \alpha\hspace{-0.2cm}\sum_{\{i,j\} \in \mathcal{E}} V_{\{i,j\}}(y_i,y_j).
\end{equation}
This is an integer optimization problem, which is NP-hard for most
interaction potentials promoting piecewise smooth segmentations.  A
remarkable exception is the binary case (when $N=2$) and submodular
interaction potentials, which \cred{are the interaction potentials that} we consider; in this
case the exact label can be computed in polynomial time by mapping the
problem onto suitable graph and computing a min-cut/max-flow on that
graph \cite{KolmogorovZ:04}.}

\parcom{Approximate solution achieved via the $\alpha$-expansion algorithm}
We find an approximate solution to this problem by using the
$\alpha$-expansion algorithm~\cite{BoykovVZ:01,BoykovK:04}.  With the
constraint that $V_{\{i,j\}}$ is metric in the label space, the local
minimum found by $\alpha$-expansion is within a known factor of the
global minimum of the labeling.

\section{Similarity Analysis}
\label{sec:similarity}
\begin{figure}[h]
  \begin{center}
    \begin{tabular}{cc}
      \includegraphics[width=.4\columnwidth]{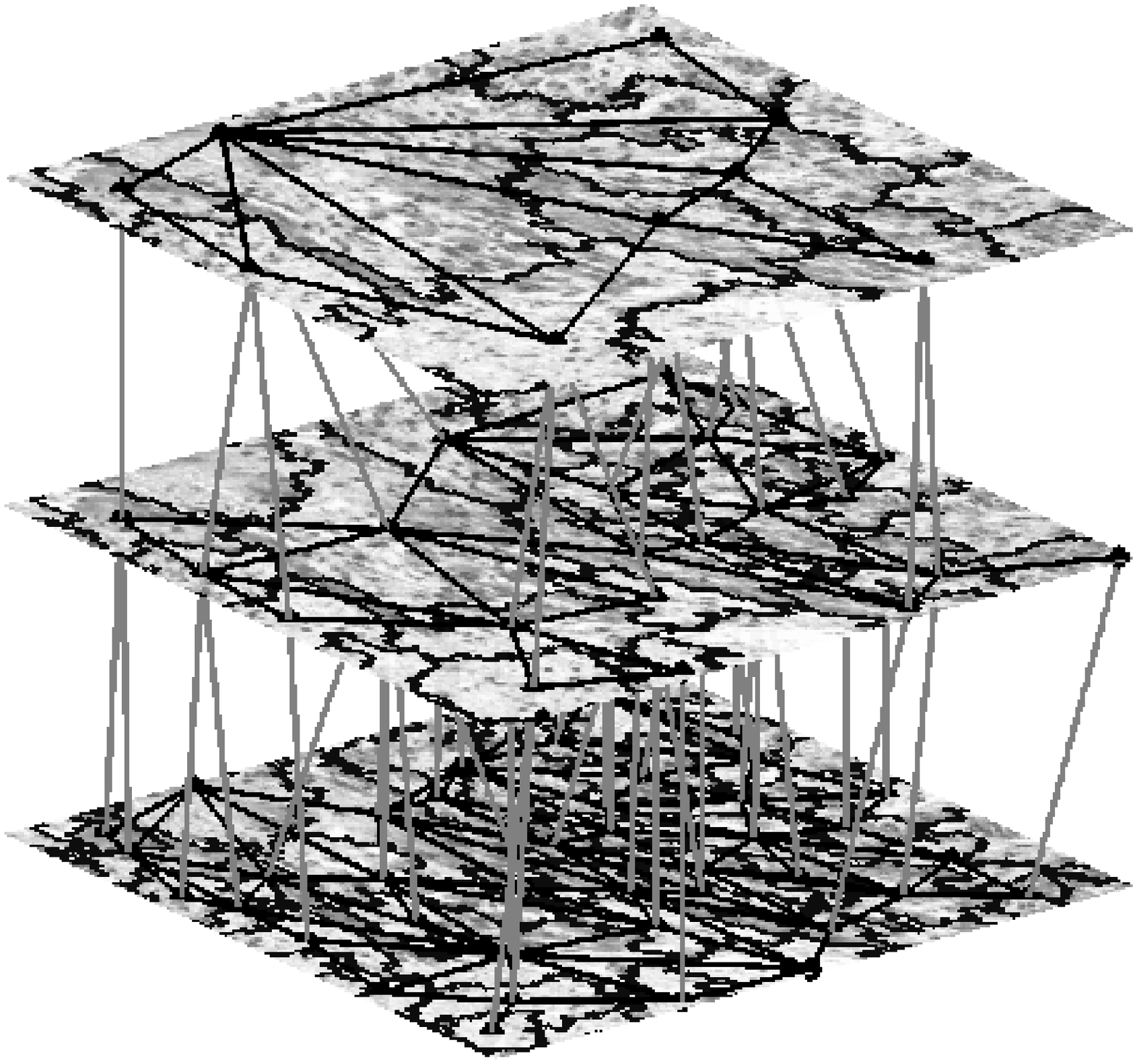} &
      \includegraphics[width=.4\columnwidth]{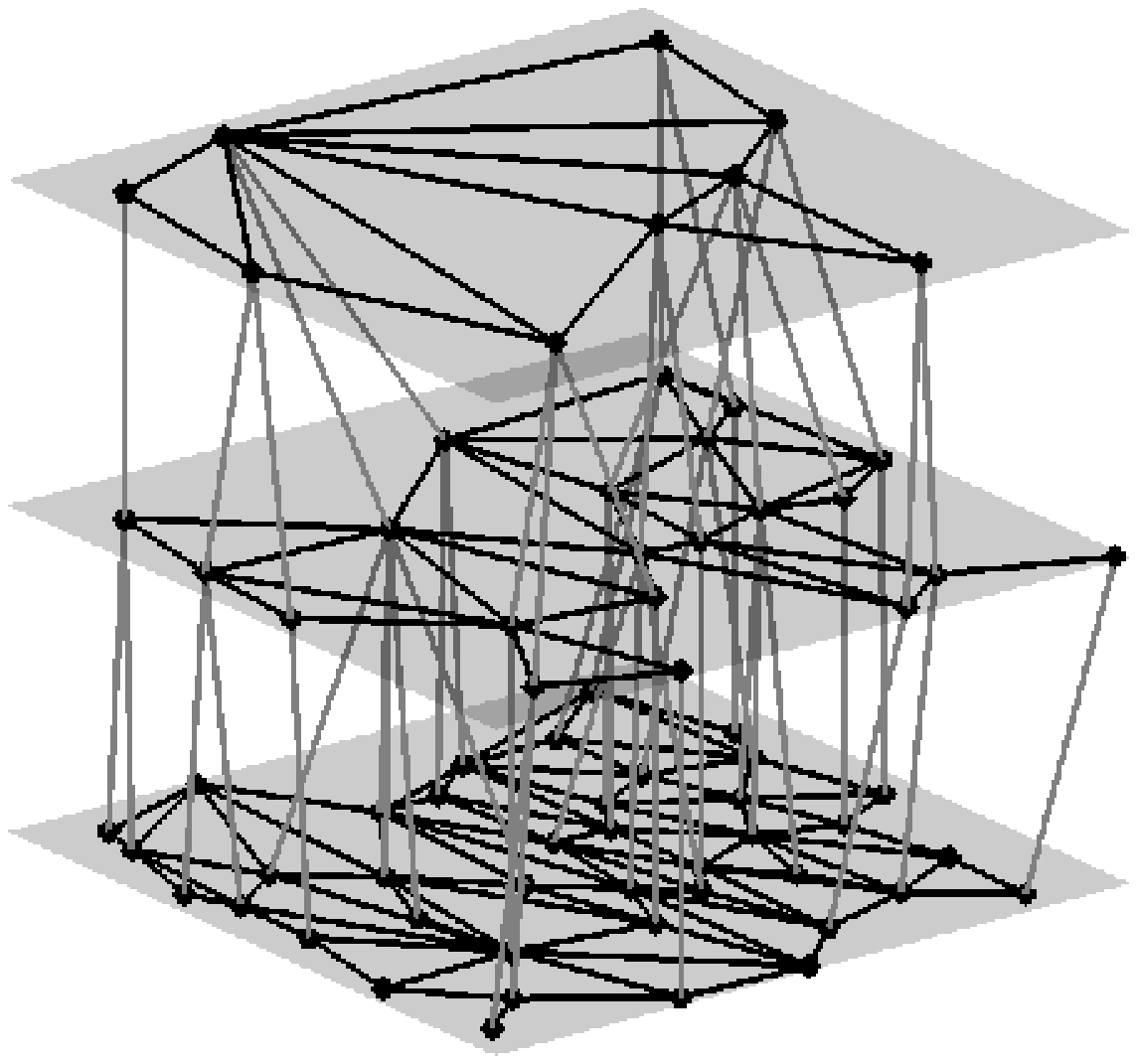}\\
      (a) Multiscale graph & (b) Multiscale graph\\
      and  multiscale partitioning & structure
    \end{tabular}
  \end{center}
   \caption{
     \label{fig:multigraph} Multiscale graph superimposed on the
     result of partitioning the same image at different scales~(a) and
     on planes denoting the different scales~(b). Nodes are denoted by
     circles, intrascale edges by gray lines, and interscale edges by
     black lines.}
 \end{figure}

\parcom{Overview of the similarity analysis}
Similarity analysis is the first step (see Fig. \ref{diag:rejclass})
of the proposed approach.  To represent similarities in the image, we
construct a similarity multiscale graph by (a) partitioning the image
at different scales and (b) finding both local and multiscale
similarities.  
\cred{The partitioning of the image at each scale is computed from the
  oversegmentation that results from using superpixels~\cite{FelzenszwalbH:04}.}
\cred{The different scales used for partitioning reflect a compromise between computational cost associated with  large multiscale graphs, and the performance gains achieved by having a multiscale graph that correctly represents the problem.}
The construction of a similarity multiscale graph \cred{(as exemplified in Fig. \ref{fig:multigraph})}
allows us to encode local similarities at the same scale, and
similarities at different scales.  The edges of the similarity
multiscale graph define the cliques present in \eqref{eq:model_DRF}.
This knowledge can be used to improve the performance of the
classification, as neighboring and similar partitions are likely to
belong to the same class.

\subsection{Multiscale Superpixels}
\parcom{Overview of the multiscale superpixelization with pixels sizes}
We obtain a multiscale partitioning of the image by computing
superpixels at different scales, that is, selecting increasing minimum
superpixel sizes (MSS)  for each superpixelization.  This leads to multiple partitions
on which the minimum number of pixels in each partition element is
changed, corresponding to a scale of the partition.
The scale selection must achieve a balance between spatial resolution and
representative partition elements (with sufficient size to compute the
statistics on the features).  

\subsection{Design of the Similarity Multiscale Graph}
\parcom{Three steps to construct the similarity multiscale graphs}
The design of the similarity multiscale graph is performed in three
steps: (1) compute a graph for each single scale partition;
(2) connect the single scale partition graphs; and (3) compute
similarity-based edge weight assignment and prune edges.  The main
idea is that a partition will have an associated graph.  By combining
partitions with different scales (an inverse relation exists between
the number of elements of a partition of an image and the scale
associated with that partition), we are able to combine graphs with
different scales.  This will be the core of the similarity multiscale
graph.

\subsubsection{Single Scale Graph as a Subgraph of the Multiscale
  Graph}
\parcom{Graph notation and obtaining a graph from a partition of
  the image}
Let us consider $P_s(I) = \bigcup_i \{\x^s_i\} $, the set of partition
elements $\x^s_i$ obtained by partitioning of the image $I$ at scale
$s$.  We associate a node $n^s_i$ to each partition element $\x^s_i \in
P_s(I)$ and defined the set of nodes at scale $s$ as
\begin{equation*}
  \mathcal{V}_s = \bigcup_i \{n^s_i\}.
\end{equation*}
There is a one-to-one correspondence between partition elements
$\x^s_i$ and nodes $n^s_i$.  For each pair of adjoint partition
elements (partition elements that share at least one pixel at their
boundary) at scale $s$, $(\x^s_i,\x^s_j)$, we create an undirected edge
between the corresponding nodes.  We have that the set of intrascale
edges at scale $s$ is
\begin{equation*}
  \mathcal{E}_s = \bigcup_i \bigcup_{j \in \mathcal{N}(n^s_i)}\{(n^s_i,n^s_j)\},
\end{equation*}
where $\mathcal{N}^s(n^s_i)$ is the set of neighbor nodes of $n^s_i$,
that is, the set of nodes that correspond to the partitions adjoint to
the partition $\x^s_i$.  Let $\mathcal{G}_s = \left(
  \mathcal{V}_s,\mathcal{E}_s \right) $ denote the graph associated to
scale $s$.  The union, for all scales, of the single scale graphs,
that is,
\begin{equation*}
  \bigcup_s \mathcal{G}_s = \bigcup_s\left(  \mathcal{V}_s,\mathcal{E}_s \right) 
\end{equation*}
is itself a graph that represents the multiscale partitioning of the
image, without edges existing between nodes at different scales.

\subsubsection{Multiscale Edge Creation}
\parcom{Connecting multiple single-scale graphs}
The multiscale graph is obtained by extending the union of all
single-scale graphs $\bigcup_s \mathcal{G}_s$ to include interscale
edges.  For $s'> s$, let $\eta(n_i^s,s')$ be a function returning a
node at scale $s'$ such that, for $j=\eta(n_i^s,s')$, we have
$\x_j^{s'}\cap \x_{n_i}^s\neq \emptyset$; that is, $j=\eta(n_i^s,s')$ is
a node at scale $s'$ whose corresponding partition element $\x_j^{s'} $
has non empty intersection with the partition element $\x_{n_i}^s$.
Based on this construction, a partition element cannot be related to
two or more different larger scale partition elements but can be
related to multiple lower level partition elements.  Let
$\mathcal{E}_{(s,s+1)}$ be the set of edges between nodes in
$\mathcal{V}_s$ and $\mathcal{V}_{s+1}$; we have that
\begin{equation*}
  \mathcal{E}_{(s,s+1)} = \bigcup_i \bigcup_{j = \eta(n_i^s,s+1)} \{(n_i^s,n_j^{s+1})\}.
\end{equation*}
The set $\mathcal{E}_{(s,s+1)}$ contains edges between adjacent
scales, connecting the finer partition at a lower scale to the coarser
partition that a higher scale.  A node at scale $s$ has exactly one
edge connecting to a node at scale $s+1$ and at least one edge
connecting to a node at scale $s-1$.

\parcom{Multiscale graph in graph notation}
Considering a set of scales $\mathcal{S}$, we have that the multiscale
graph $\mathcal{G}$ resulting from the multiscale partitioning is
\begin{equation*}
  \mathcal{G} = \left(\underbrace{\bigcup_{s=1}^{|\mathcal{S}|} \mathcal{V}_s}_\textrm{nodes},\underbrace{\left( \bigcup_{s=1}^{|\mathcal{S}|} \mathcal{E}_s\right)}_\textrm{intrascale edges} \cup \underbrace{\left(\bigcup_{s=1}^{|\mathcal{S}|-1} \mathcal{E}_{(s,s+1)}\right)}_\textrm{interscale edges}\right) = \left( \mathcal{V},\mathcal{E}\right).
\end{equation*}

\subsubsection{Edge Weight Assignment}
\parcom{Edge assignement and edge prunning in the multiscale graph}
Given the multiscale graph $\mathcal{G}$, we now compute and assign
edge weights based on similarity.  Let $f_\textrm{si}(n_i^s)$ be a
function that computes similarity features on the node $n_i^s$,
corresponding to the partition element $\x_i^s$.  The weight of the
edge $(n_i^s,n_j^{s'}) \in \mathcal{E}$ is computed as
\begin{equation}
  \label{eq:edge_weight}
  w_{n_i^s,n_j^{s'}} \propto v(s,s') \exp{(- \| f_\textrm{si}(n_i^s) - f_\textrm{si}(n_j^{s'})\|^2 / \gamma)},
\end{equation}
where $\gamma$ is a scale parameter, $\exp{(- \| f_\textrm{si}(n_i^s)
  - f_\textrm{si}(n_j^{s'})\|^2 / \gamma)}$ quantifies the similarity
between two nodes $n_i^s$ and $v(s,s')= v_{\text{intrascale}}$, if
$s=s'$, and $v(s,s') = v_{\text{intercale}}$, if $s\neq s'$.  The
rationale for different weights for intrascale and interscale edges
comes from the different effect of the multiscale structure.  For a
given value of intrascale weight, lower values of the interscale edge
weight downplay the multiscale effect on the graph, and higher values
of the interscale edge weight accentuate the multiscale effect.

\section{Expert Classification}
\label{sec:classification}
\parcom{Expert classification overview}

The expert classification block of the system is \cred{constructed from} two sequential steps: feature extraction and classification.
The feature extraction step consists in computing the application-specific features and extracting statistics of the features on each of the lowest level partitions.
In the classification step, the classifier is trained, applied to the data, and the classification risk is computed.
\cred{
As the feature extraction procedure was introduced in Section \ref{subsec:features} and is application-dependent, and the classification procedure was described in Section \ref{subsec:classification}, we will focus on the computation of the classification risk.
}

\subsection{Rejection by Risk Minimization}
\parcom{Rationalle for rejection using risk minimization}
By \cred{approaching} classification as a risk
minimization problem, we are able to introduce rejection.  To improve
accuracy at the expense of not classifying all  partitions, we
classify while rejecting.  Let $\L' = \L \cup \{N+1\}$ be an extended
set of partition class labels with an extra label.  
\cred{The rejection class can be considered as an \emph{unknown} class that represents the inability of the classifier to correctly classify all samples.
The extra label $N+1$ corresponds to this rejection class.
}


\subsubsection{Classification with Rejection by Risk Minimization}
\parcom{Rejection via risk minimization: top-level overview}
Given a feature vector $\f_{i}$, associated to a partition element
$\x_i$, and the respective (unobserved) label $y_i\in\cal L$, the
objective of the proposed classification with rejection is to estimate
$y_i$, if the estimation is reliable, and do nothing (rejection)
otherwise.

\parcom{Risk minimization: the cost matrix}
To formalize the classification with rejection, we introduce the
random variable $\hat{y}_i \in {\cal L}'$, for $i\in \cal V$, where
$\hat{y}_i = N + 1 $ denotes rejection. In addition, let us define a
$(N+1) \times N$ cost matrix $C = [c_{j_1,j_2}]$ where the element
$c_{j_1,j_2}$ denotes the cost of deciding that $\hat{y} _i= j_1$,
when we have $y_i= j_2$ and does not depend on $i\in\cal V$.

\parcom{Risk minimization: definition}
Let the classification risk of $\hat{y}_i = k$
conditioned to $\f_{i}$ be defined as:
\begin{align*}
  R(\hat{y}_i = k|\f_{i})  & = \mathbb{E}_{y_i}[c( \hat{y}_i = k, y_i )|\f_{i}] \\
  & = \sum_{j_2=1}^{N} c_{k,j_2} p(y_i=j_2|\f_{i},\widehat{\W}).
\end{align*}
By setting  $c_{N+1,j_2} = \rho$,  we get
\begin{align}
  \label{eq:risk_definition}
  R(\hat{y}_i = k, k \neq N+1|\f_{i})  = & \sum_{j_2=1}^{N} c_{k,j_2} p(y_m=j_2|\f_{i},\widehat{\W}), \nonumber  \\
  R(\hat{y}_i = k, k = N+1|\f_{i}) = & \rho.
\end{align}
By minimizing
\eqref{eq:risk_definition} over all possible partition labelings
$\L'^{| \S |}$, we obtain
\begin{align}
  \widehat{y} \ =& \  \arg\min_{y\in\L'^{ |\V|}} \sum_{i\in\V}  R(y_i |  \f_{i}).
\label{eq:rej_minimization}
\end{align}
Note that if $c_{j_1,j_2} = 1-\delta_{j_1 - j_2}$, where $\delta_{n}$
is the
Kronecker delta function, minimizing~\eqref{eq:rej_minimization}
yields
\begin{displaymath}
  \hat{y_i} \ = \ \left\{
    \begin{array}{rl}
      \displaystyle{\arg\max_{y_i \in \L} p(y_i|\f_{i},\widehat{\W})}, &
      \displaystyle{\max_{y_i \in \L}  p(y_i|\f_{i},\widehat{\W}) > 1-\rho};\\
      N+1 ,& \mbox{otherwise}.
    \end{array}
  \right.
\end{displaymath}
In other words, if the maximum element of the estimate of the
probability vector is large, we are reasonably sure of our decision
and assign the label as the index of the element; otherwise, we are
uncertain and thus assign the unknown-class label.

\subsubsection{Including Expert Knowledge}
\parcom{Expert Knowledge for risk minimization: rationalle}
Expert knowledge can be included in the risk minimization.  Class
labels can be grouped in $L$ superclasses $\mathcal{L} =
\{\mathcal{L}_1 , \hdots , \mathcal{L}_L \}$ (\cred{each super class is an element of the} partition of the set
of classes $\mathcal{L}$) on which misclassification within the same
superclass 
 should
have a cost different than misclassifications within different
superclasses.

\parcom{Risk minimization with expert knowledge: definition}
Let us now consider the following cost elements with a cost $g$ for
misclassification within the same superclass,
\begin{equation*}
  c'_{j_1,j_2} =
  \begin{cases}
    0 & \mbox{if } j_1 = j_2;\\
    g & \mbox{if $j_1$ and $j_2$ belong to the same superclass};\\
    1 & \mbox{otherwise}.
  \end{cases}
\end{equation*}
The expected risk considering expert knowledge of selecting the class
label $y_i \in \L'$ in the partition is
\begin{align}
  \label{eq:exp_risk_definition}
  R'(\hat{y}_i = k,k \neq N+1 |\f_{i} )  =  &     \sum_{j_2=1}^{N} c'_{k,j_2} p(y_m=j_2| \f_{i},\widehat{\W}) , \nonumber \\
  R'(\hat{y}_i = k,k = N+1 |\f_{i})  = & \rho .
\end{align}
Minimizing \eqref{eq:exp_risk_definition} over all possible partition labelings yields
\begin{equation*}
  \yhat'_i \ = \  \arg\min_{y\in\L'^{ |\V|}} \sum_{i\in\V}  R'(y_i \mid \f_{i}, \widehat{\W}).
  \label{eq:exp_rej_minimization}
\end{equation*}
This formulation allows us to include expert knowledge in the
assessment of a risk of assigning a label.

\section{Contextual Rejection}
\label{sec:contextual_rejection} 
\subsection{Problem Formulation}
\parcom{Contextual rejection formulated as risk minimization over a
similarity graph}
We formulate the problem of classification with rejection using
contextual information as a risk minimization problem defined over the
similarity multiscale graph $\mathcal{G}$.

\parcom{Posing the problem as an optimization problem}
As shown in \eqref{eq:class_np_1}, we can pose the classification problem
as an energy minimization problem of two potentials over the
undirected graph $\mathcal{G} = (\mathcal{V},\mathcal{E})$
representing the multiscale partitioning of the image $I$.  The
association potential $D$ is the data term, the interaction potential
$V_{\{i,j\}}$, for $(i,j) \in \mathcal{E}$, is the contextual term,
and $\alpha \in [0,1]$ is a weight factor that balances the relative
weight between the two is denoted as contextual index. Then,
\begin{equation}
  \label{eq:problem_formulation}
  \yhat = \arg\min_{y \in \mathcal{L}'^{|\V|}} (1-\alpha) \sum_{i \in \mathcal{V}} D(y_i, \f_i) + \alpha \sum_{(i,j)\in \mathcal{E}} V_{\{i,j\}}(y_i,y_j).
\end{equation}


\subsection{Association Potential: Expert Knowledge}
\parcom{Association potential as classification risk}
The association potential measures the disagreement between the
labeling and the data; we formulate it as \cred{a strictly increasing function of} the classification risk in \eqref{eq:exp_risk_definition}:
\begin{equation*}
  D(y_i,\f_i) =   \cred{\log}(R'(y_i \mid  \f_{i}, \What)), \mbox{ for }  i \in \mathcal{V}.
\end{equation*}
\cred{This unary association} potential is associated with the nodes $\mathcal{V}$ of the
graph (partitions), and includes the rejection \cred{ that is} present in the classification risk $R'$.

\subsection{Interaction Potential: Similarity }
\parcom{Similarity as interaction potential: overview}
The {interaction potential is based on the topology of the graph
$\mathcal{G}$, combining intra and inter level interactions between
the pairs of nodes connected by edges, based on their similarity.  We
define an interaction function $\psi$ that enforces piecewise smooth
labeling among the pairs of nodes connected by edges.

\parcom{Encoding the difference between intra and inter level edges
in the graph}
In the design of the similarity multiscale graph, the difference
between intralevel and interlevel edges is encoded in different
multiplier constants of the edge weight \eqref{eq:edge_weight}.  This
allows us to work with intralevel and interlevel edges in the same
way, without increasing the complexity of the pairwise
potential.  Accordingly, we set
\begin{equation*}
  V_{\{i,j\}}(y_i,y_j) =   w_{i,j} \psi(y_i,y_j),
\end{equation*}
where $w_{i,j}$, for $(i,j) \in \mathcal{E}$, corresponds to the edge
weight defined in \eqref{eq:edge_weight}.

\subsubsection{Interaction function}
\parcom{Simple interaction function}
The interaction function $\psi$ enforces piecewise smoothness in
neighboring partitions; its general form is $\psi(y_i,y_j) = 1 -
\delta_{y_i-y_j}$, that is $0$ if $y_i = y_j$ and 1 otherwise.

\parcom{Complex interaction function}
It is desirable, however, both to ease the transition into and out of
the rejection class, and ease the transitions between classes belonging to
the same superclass.  We achieve this by adding a superclass
consistency parameter $\psi_C$ and a rejection consistency parameter
$\psi_R$ to the interaction potential as follows:
\begin{equation}
  \psi(y_i,y_j) = \begin{cases}
    0 & \mbox{if } y_i = y_j;\\
    \psi_C & \mbox{if $y_i$ and $y_j$ belong to the same superclass};\\
    \psi_R & \mbox{if $y_i = N+1$ or $y_j = N+1$};\\
    1 & \mbox{otherwise}.
\end{cases}
\end{equation}

\parcom{Theoretical guarantees of the complex interaction
  function. Connection with rejection structure}
Defining a rejection consistency parameter $\psi_R$ allows us to have
an interaction function that can be metric, meaning that the
interaction potential will be metric.  Another effect is the ability
of controlling the structure of the rejected area.  With a rejection
consistency parameter close to $0$ we  obtain a labeling with
structure with unstructured rejection; this means that rejection areas
can be spread on the image and can consist of one partition element
only.  With a higher value, we are imposing structure both on the
labeling but also on the rejection areas, leading to larger and more
compact rejection areas.

\section{Experimental Results}
\label{sec:experimental}
\cred{
\parcom{Overview of the experimental results}
With the framework for image classification with rejection using
contextual information in place, we will now show examples of its application in real data.
The main applicational area of the framework is tied with a subclass
of image classification problems described in the introduction: ill-posed classification problems where the access to representative pixelwise ground truth is prohibitive; the pixels can belong to uninteresting or unknown classes; and the need for thigh accuracy surpasses the need to classify all samples.

The first example, the classification of natural images (Section~\ref{subsec:natural}), illustrates the generality of the framework.
Whereas designed for a subclass of image classification problems, the proposed framework can also be applied to more general image classification problems: supervised segmentation of natural images.
The second example, the classification of H\&E stained teratoma tissue images (Section~\ref{subsec:HE}), shows the advantages of using a robust classification scheme combining rejection and context on the main applicational area of this framework. 

With the classification of natural images, we also explore the effect of the graph structure on the classification of an image: how the classification with rejection propagates through the different layers of the multiscale graph; and how the number of scales, or ``depth'' of the multiscale graph, affects the performance of the classification.
With the classification of H\&E images, we also explore the joint interaction between context and rejection in the classification problem, and the behavior of the framework as the difficulty of the classification problem increases.

As the concept of combining classification with context with classification with rejection in pixelwise image classification is novel, there are no competing methods nor frameworks to compare to.
To provide an assessment of the performance of the framework, we
compare the performance of the framework with the performance of
context only, and with the performance of rejection only, with
selection of optimal rejected fractions.


}
\afterpage{\clearpage}
\subsection{Natural Images}
\label{subsec:natural}
\parcom{Illustrating the method on horses}
We illustrate the flexibility of the formulation, by applying the formulation \cred{to} the classification of natural images (Fig. \ref{fig:natural}).
We obtain a multilevel classification from an image extracted from the BSD500 data set~\cite{BSD500}. 
\subsubsection{\cred{ Experimental setup}}
Both the application-specific features (for classification) and the
similarity features (for graph construction) are the \cred{color on
  the RGB colorspace, and the statistic extracted from the partition
  elements is the sample mean of the RGB color space inside the partition element.
This means that, for the $i$th partition element $\x_i$, both the application-specific and the similarity features consist of the sample mean of $\{\z_j,  j \in \x_i\}$, the RGB color space inside the partition element.}
\cred{
The number of classes is $K=3$,
where $10$ randomly selected superpixels from the lowest scale are
used to train the classifier. No superclass structure is assumed.
}

\subsubsection{\cred{ Effect of the multiscale graph on the classification}}
\cred{The effect of the multiscale graph on the classification is illustrated on Fig. \ref{fig:natural}: finner segmentations on the smaller scales, with disjoint rejected areas; and coarser segmentations on the larger scales with a large rejection area.
Due to the characteristics of the superpixelization, the class boundaries appear natural in all scales.

\begin{figure}[ht]
  \begin{center}
    \begin{tabular}{cccc}
      \includegraphics[width=.225\columnwidth]{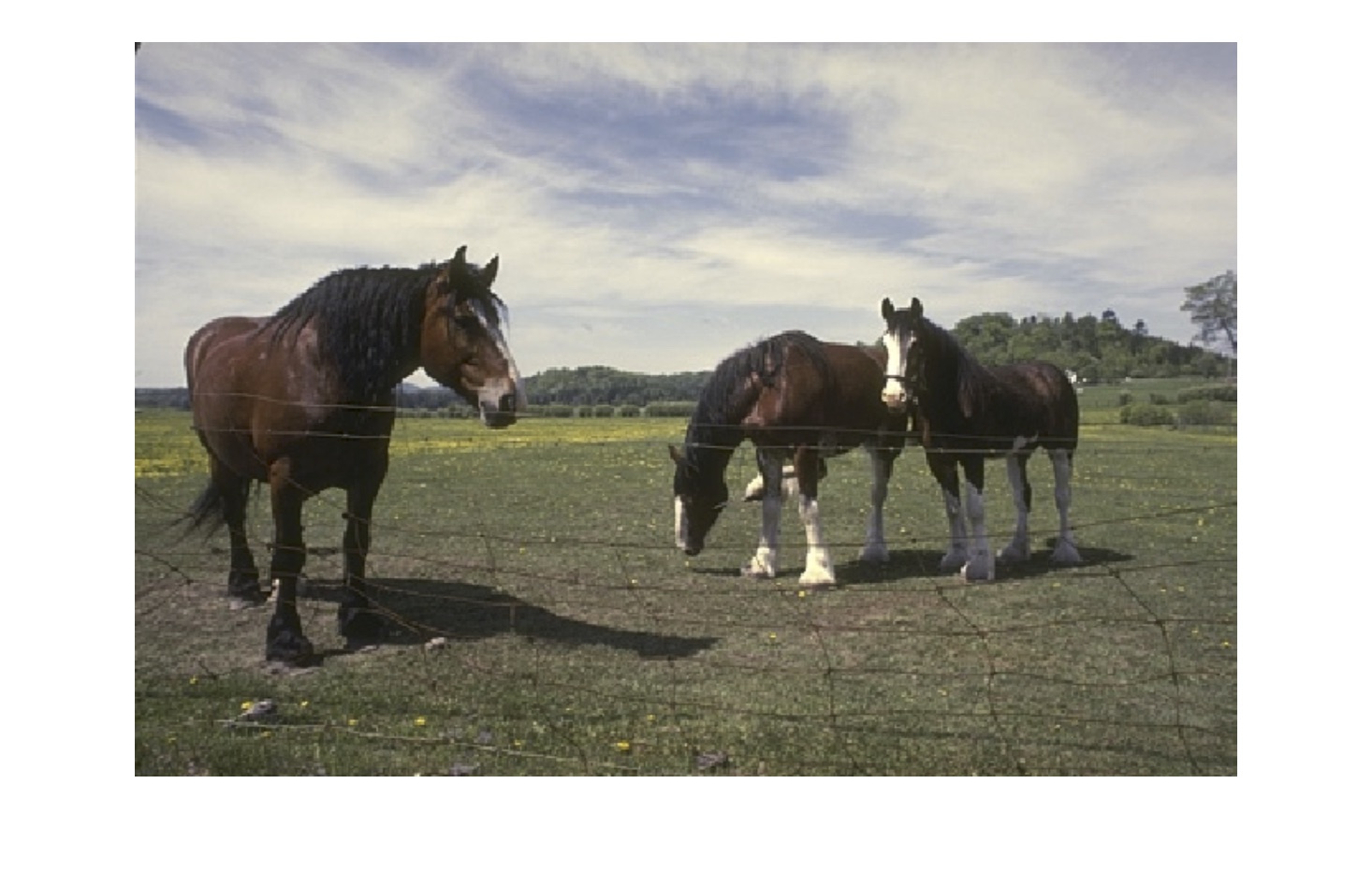} & 
      \includegraphics[width=.225\columnwidth]{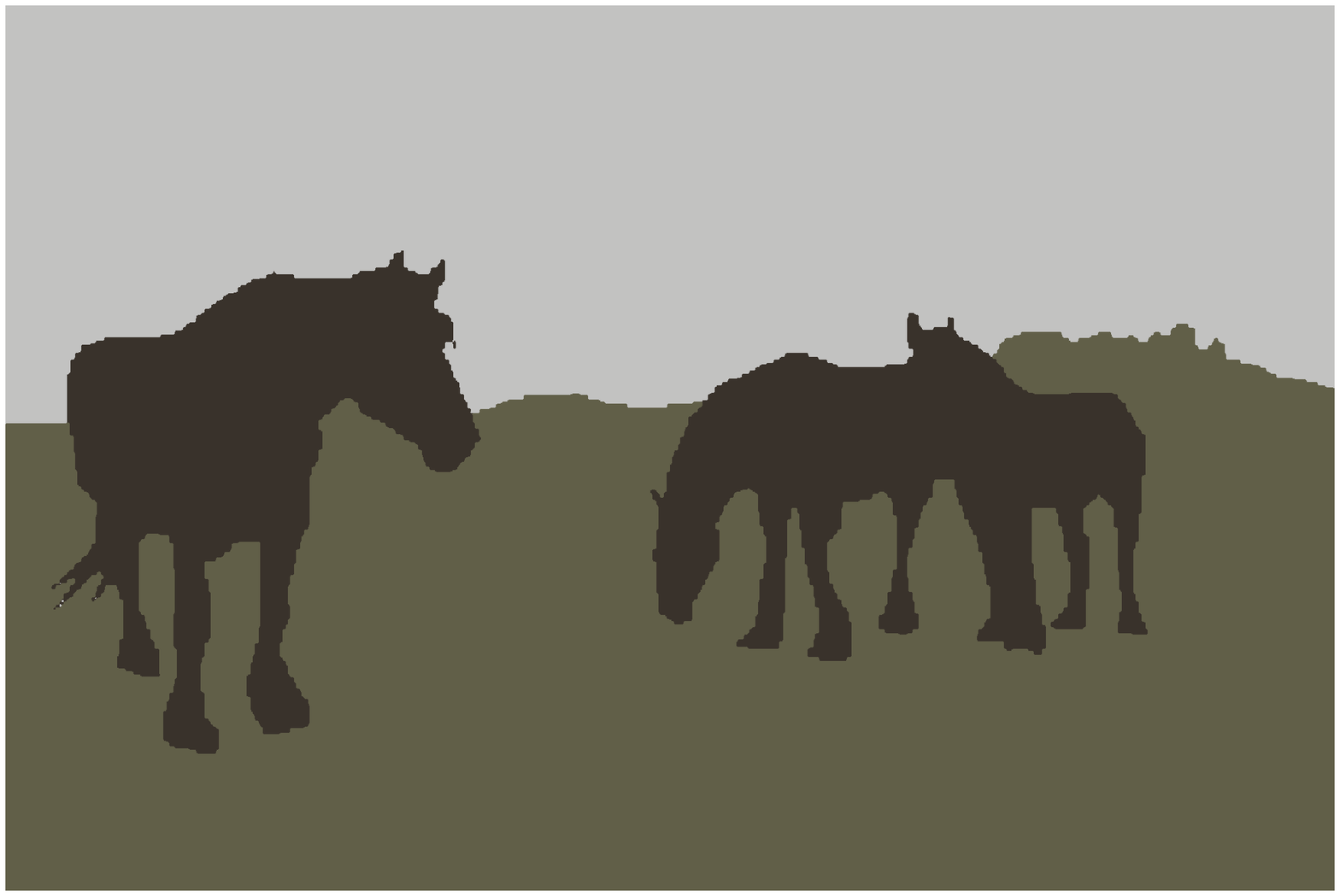}&
      \includegraphics[width=.225\columnwidth]{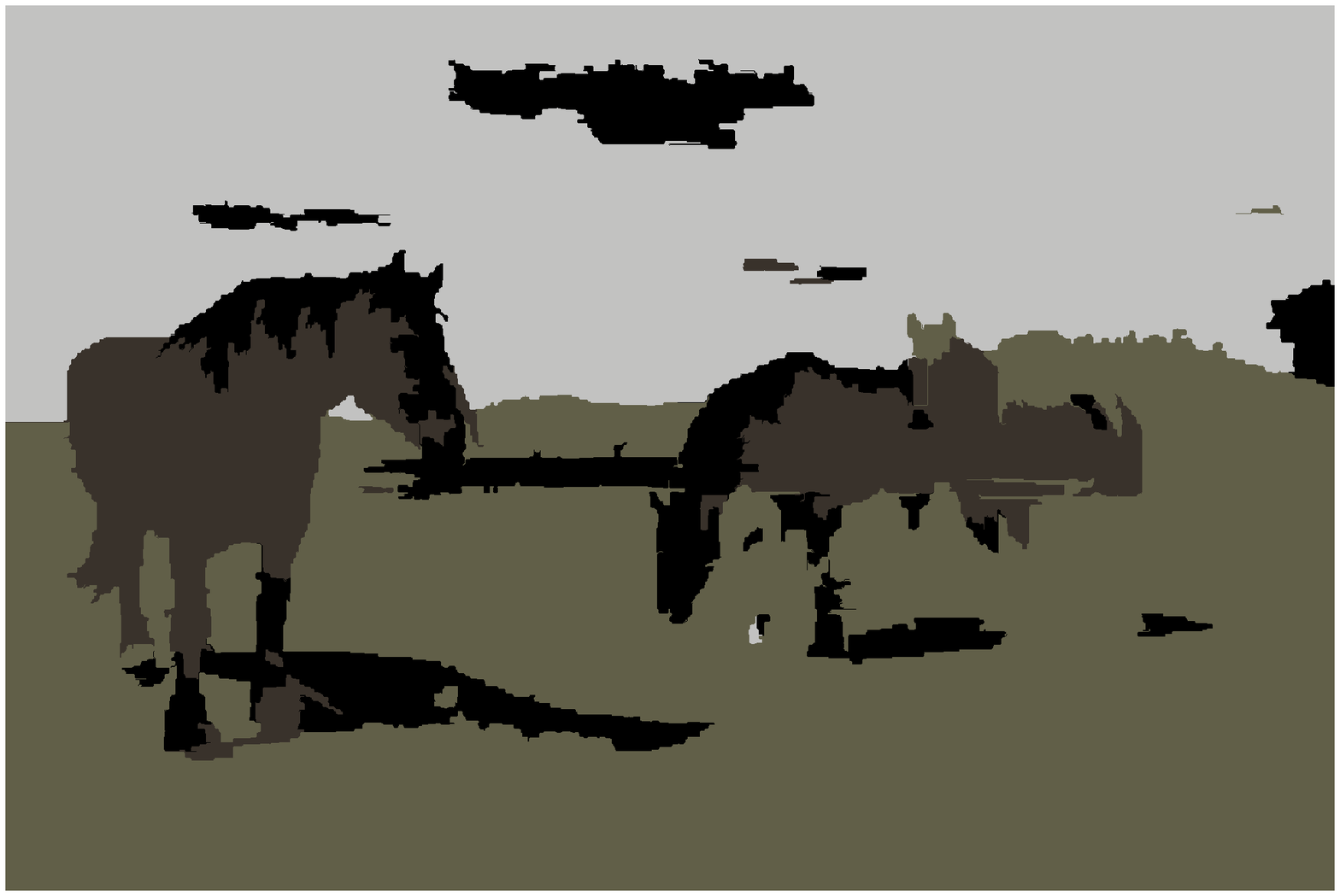} &
      \includegraphics[width=.225\columnwidth]{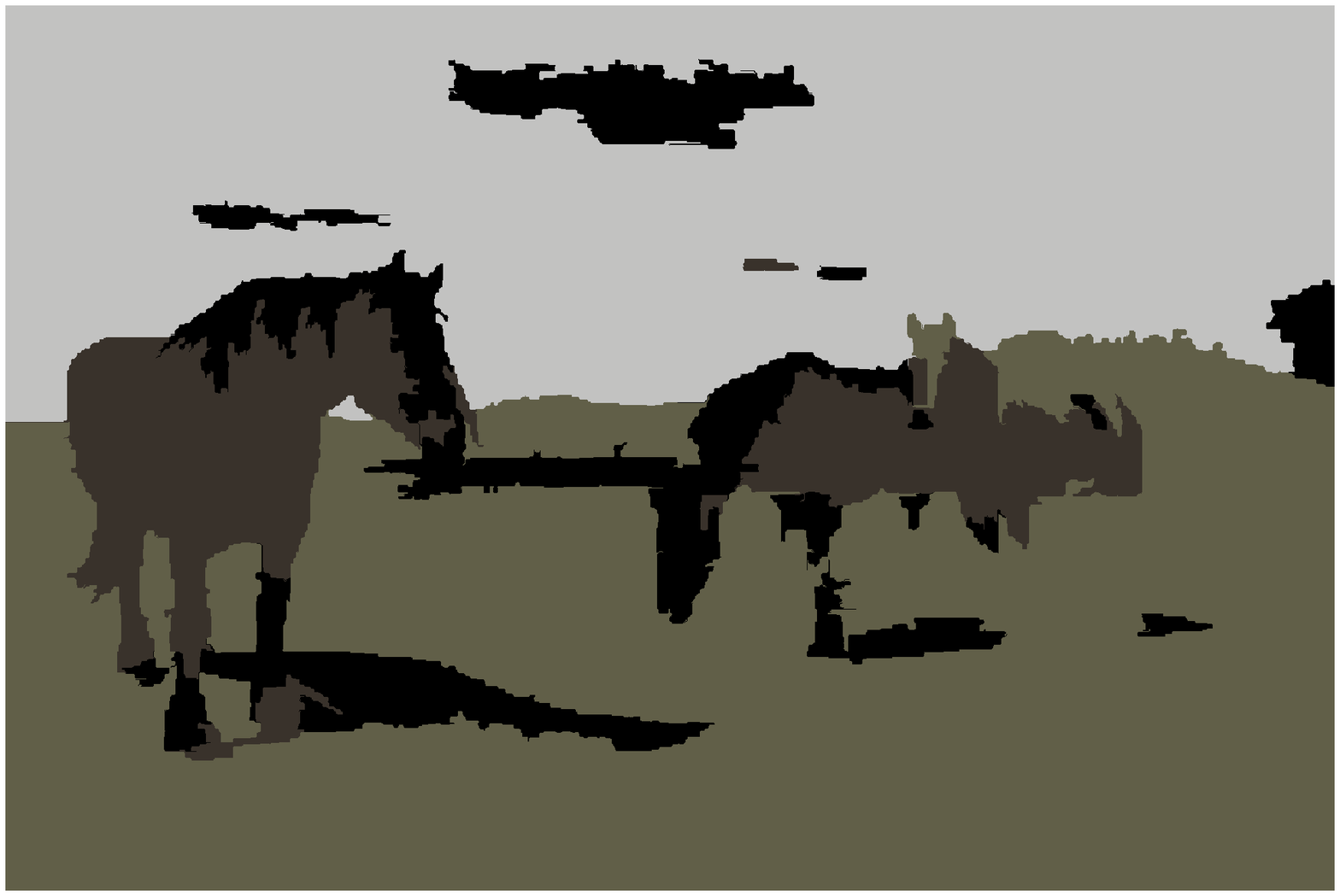} \\
      Original image  & Ground truth &      Level $s= 1$ & Level $s= 2$ \\
      \includegraphics[width=.225\columnwidth]{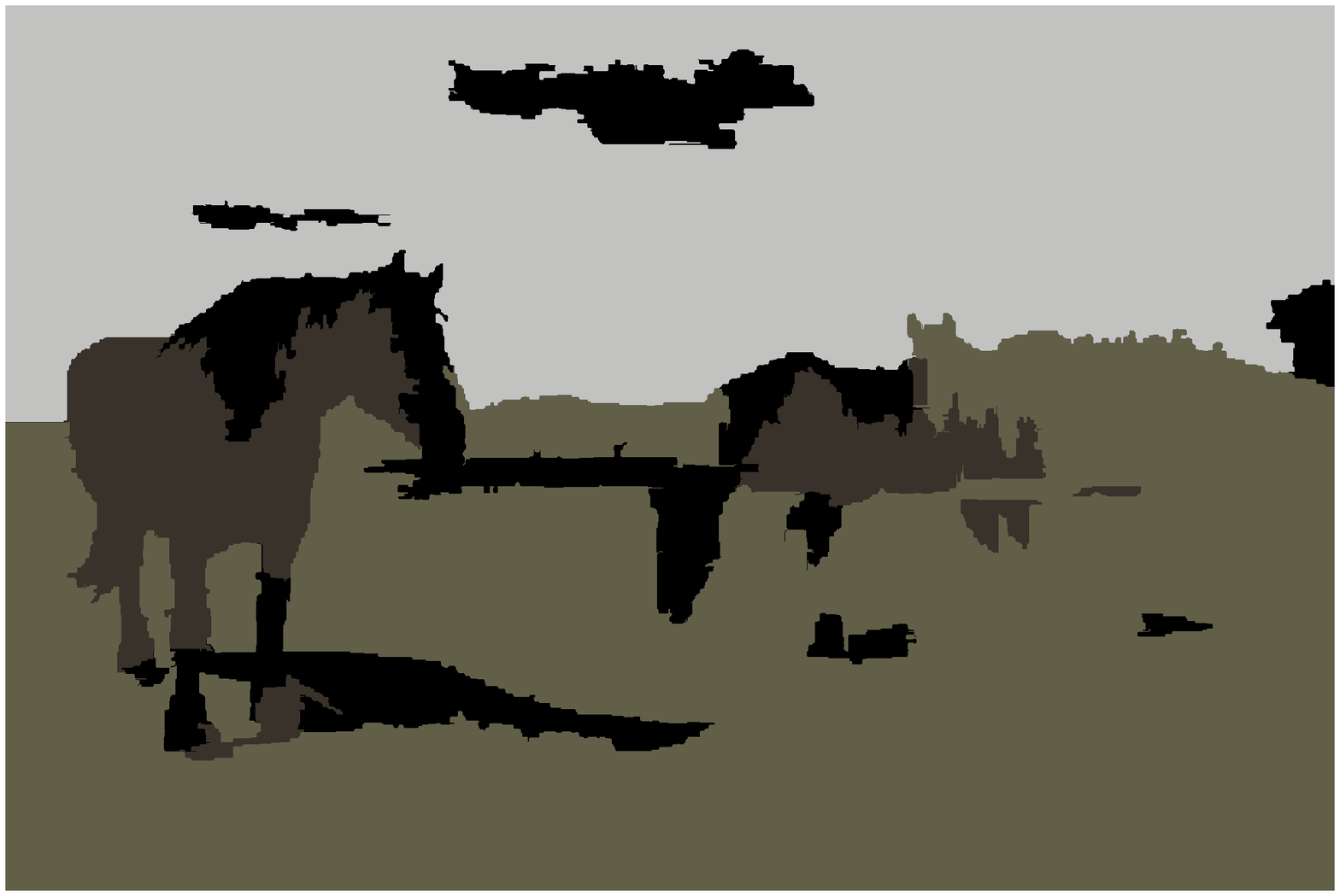}&
     \includegraphics[width=.225\columnwidth]{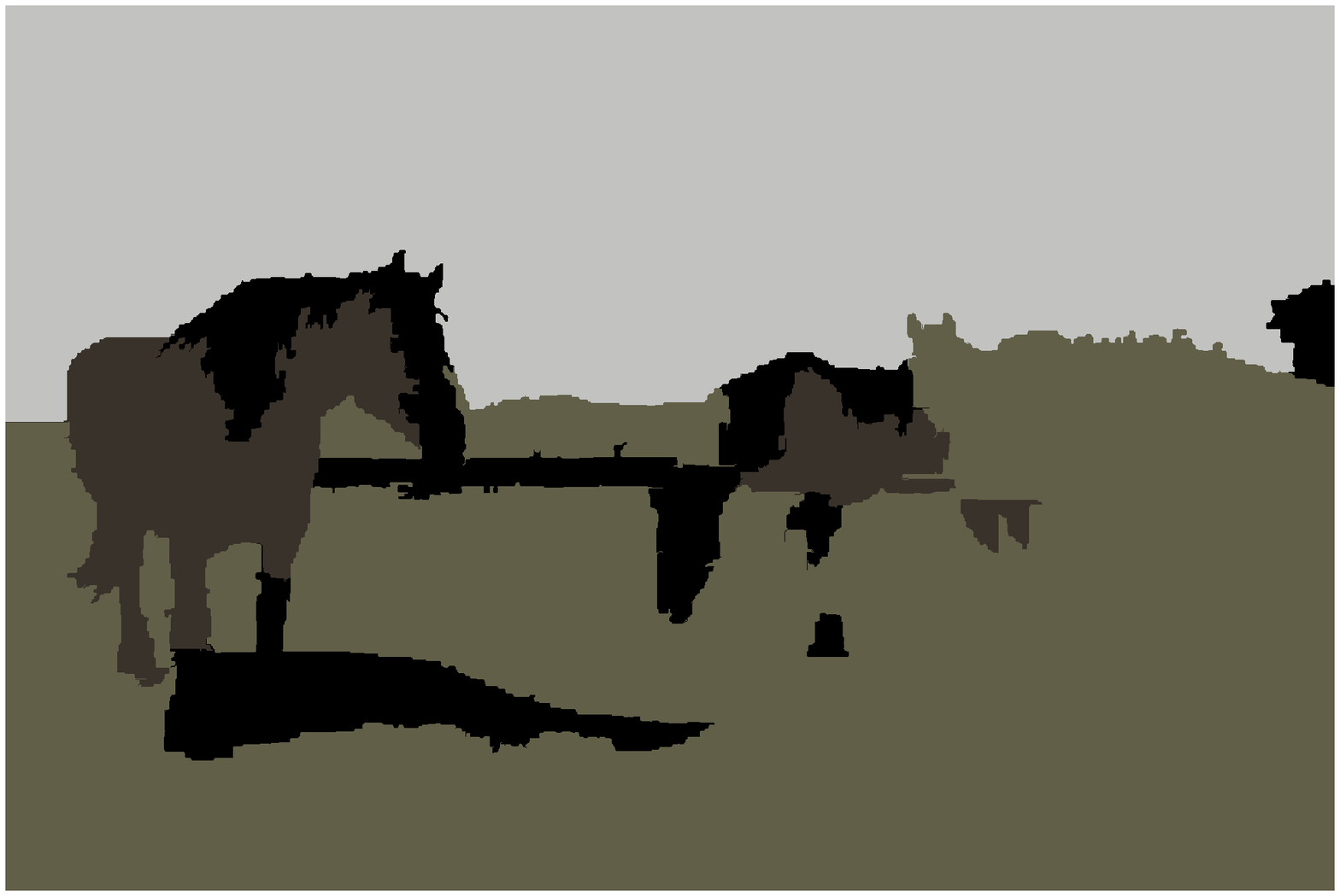} &
      \includegraphics[width=.225\columnwidth]{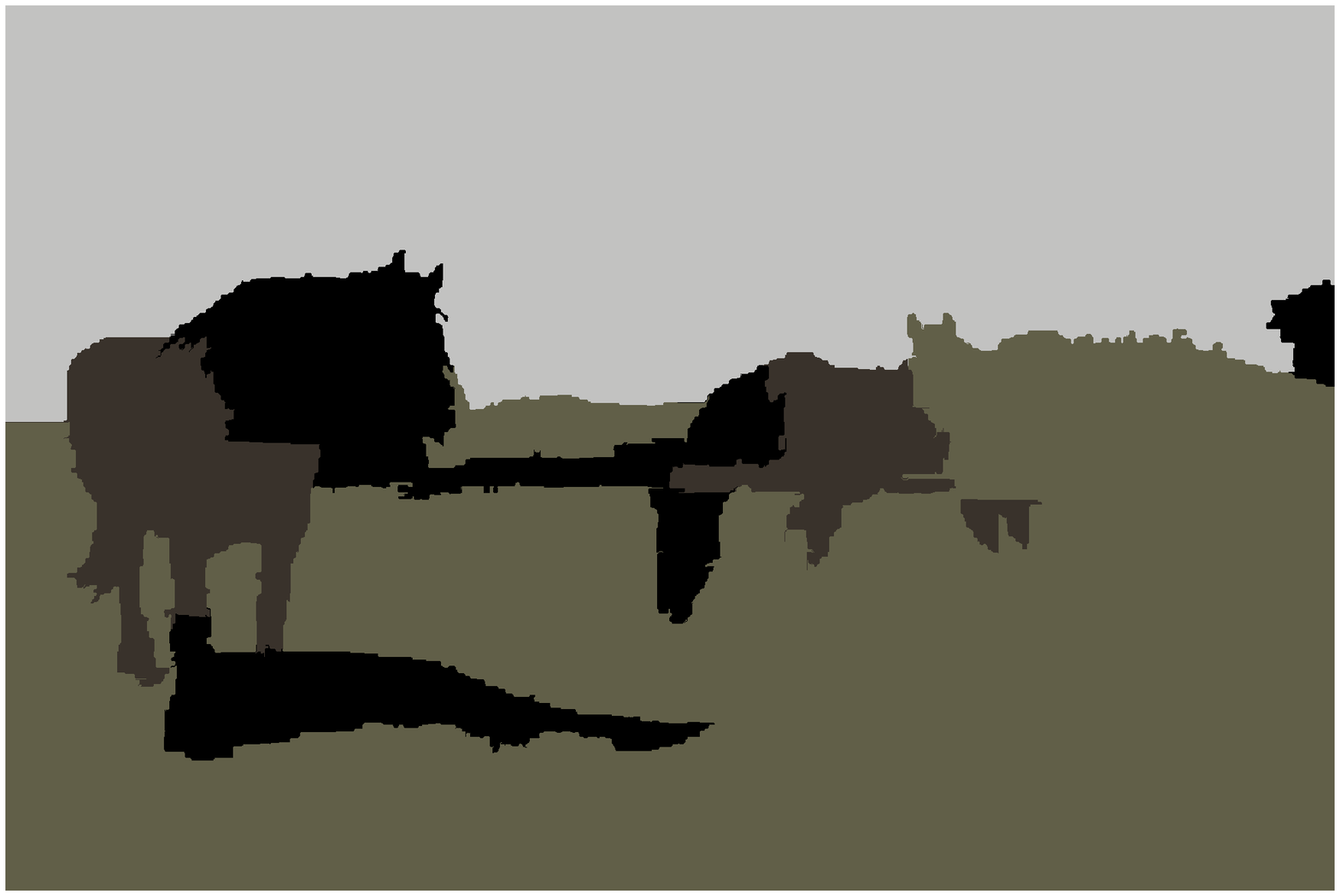} &
      \includegraphics[width=.225\columnwidth]{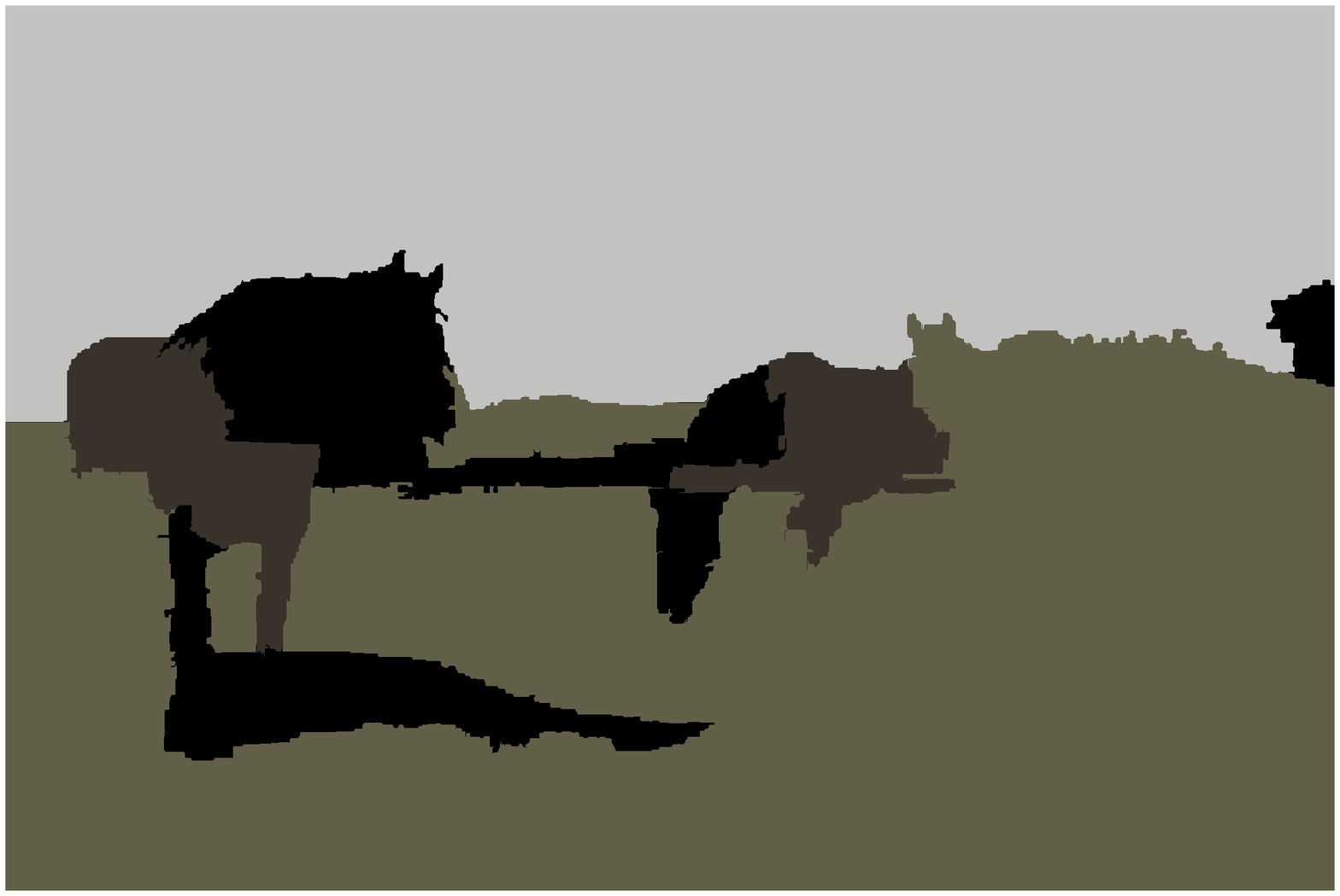}\\
      Level $s= 3$ & Level $s= 4$ &      Level $s= 5$ & Level $s= 6$ \\
\end{tabular} 
\caption{\label{fig:natural} Example of classification  with
  rejection (in black) across  multiple levels in a natural image
  from the BSD500 data set. }
  \end{center}
\end{figure}

We illustrate the robustness of the framework with regard to the number of scales by comparing the classification performance with a varying number of scales (Fig. \ref{fig:scale}).
The variation of the number of scales is achieved by stacking coarser
single-scale graphs on the multiscale graph, through an increase of
the minimum superpixel size (MSS) by a factor of $2$: $1$ scale
corresponds to a single scale graph of MSS $100$, $2$ scales to a
multiscale graph of MSS of $(100, 200)$, up to $11$ scales, that
corresponds to a multiscale graph of MSS of $(100, 200, \hdots, 6400)$.

In Fig. \ref{fig:scale} it is clear the performance improvement of using multiscale similarity graphs (more than one scale) against single scale similarity graphs (just one scale).
The stabilization of the mean performance for more than $4$ scales is an indicator of the robustness of the framework with regard to the number of scales.
}

\begin{figure}[htb]
  \begin{center}
    \begin{tabular}{cc}
\includegraphics[width=.5\columnwidth]{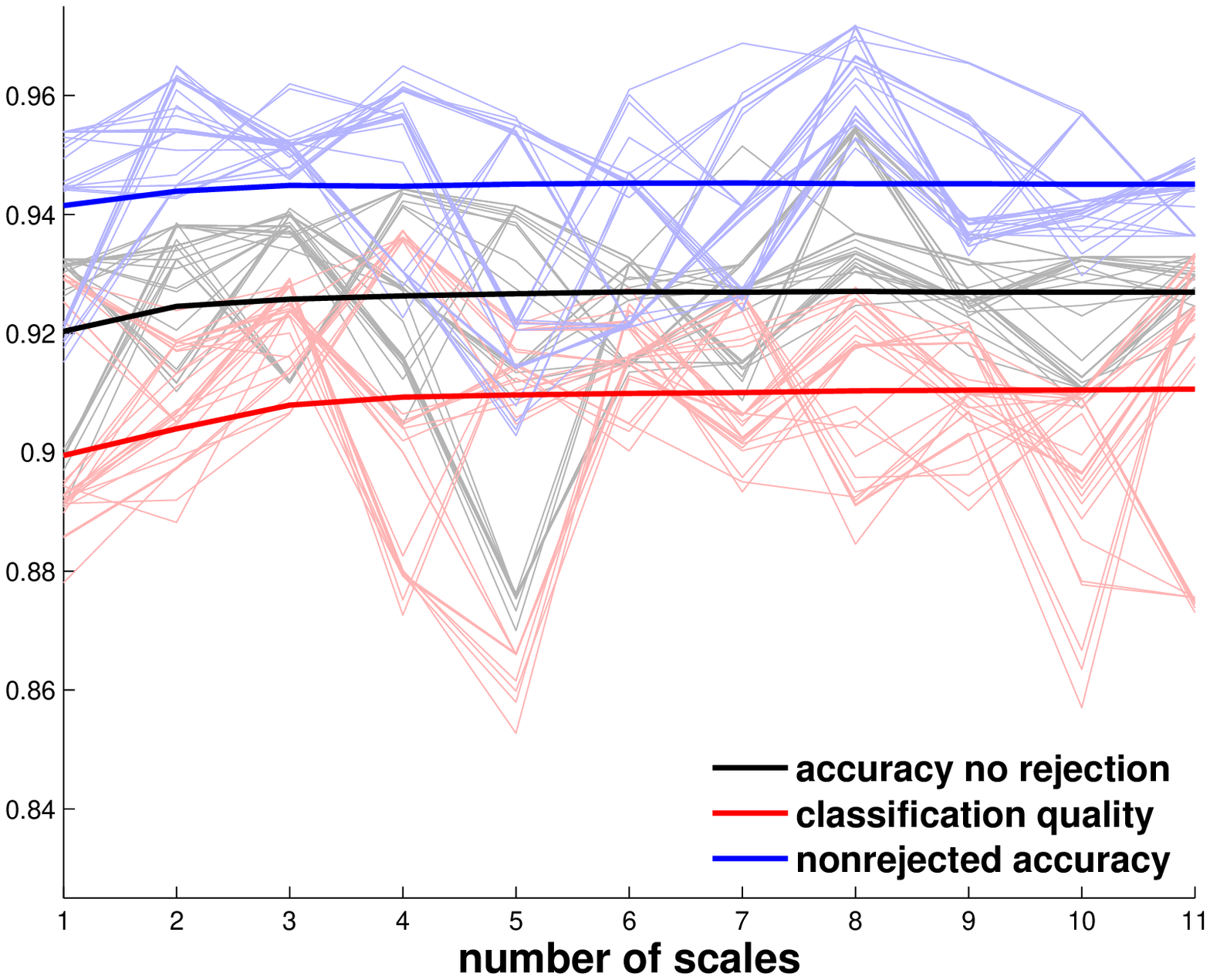} &
\includegraphics[width=.5\columnwidth]{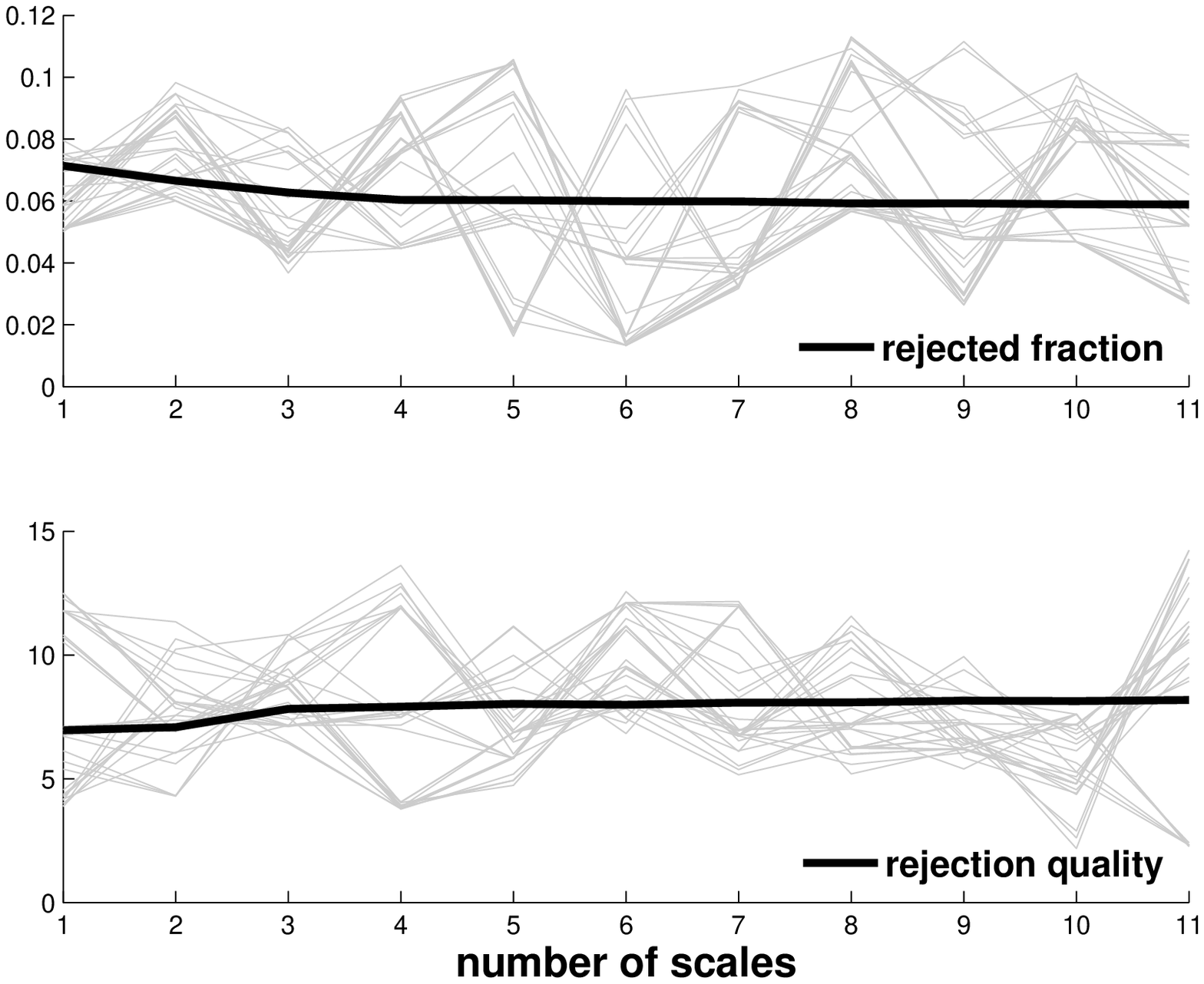}\\
\end{tabular}
\caption{\label{fig:scale}  \cred{Evolution of classification
    performance with number of scales. Results obtained from $30$
    Monte Carlo runs with    different training sets of $10$ randomly selected samples per
    class of the image in Fig. \ref{fig:natural}. The variation of performance for more than $4$ scales is negligible. }}
  \end{center}
\end{figure}
\afterpage{\clearpage}
\subsection{H\&E Data Set}
\label{subsec:HE}
\parcom{H\&E data set description}
Our H\&E data set consists of 36 $1600\times1200$-size images of H\&E
stained teratoma tissue slides imaged at $40\times$ magnification
containing $20$ classes; Fig. \ref{fig:class_res} shows three examples.

\cred{
\subsubsection{Experimental Setup}
As application-specific features we use the \emph{histopathology vocabulary (HV)}~\cite{BhagavatulaFKGOCK:10,BhagavatulaMFCOK:13}.
These features emulate the visual cues used by expert histopathologists~\cite{BhagavatulaFKGOCK:10,McCannBYFOK:12,BhagavatulaMFCOK:13}, and are thus physiologically relevant.  From the HV, we use nucleus size (1D), nucleus eccentricity (1D), nucleus density (1D), nucleus color (3D), red blood cell coverage (1D), and background color (3D).
As similarity features we use the color on the RGB colorspace.

The statistic extracted for the application-specific and the similarity features, on the lowest level
of the partition, consists of the sample mean of the feature values on the partition. It is a balance between good classification performance, low feature dimensionality, and low complexity.
This results in $10$ dimensional application-specific feature vectors, and $3$ dimensional similarity feature vectors.
The superclasses are constructed from the germ layer (endoderm, mesoderm, and ectoderm).
Classes derived from the same germ layer will belong to the same
superclass.

The multiscale similarity graph is built with six scales with a MSS of
$(100, 200, 400, 800, 1600, 3200)$ for each of the layers of the
similarity graph.
This provides a compromise between the computational burden associated
with large similarity graphs and the performance increase obtained.
The results we present with six scales are marginally better than the
ones achieved with five or seven scales.

}
\subsubsection{Parameter Analysis}
\parcom{Overview of parameter analysis}
In this section we analyze the impact of regularization parameter
$\lambda$ on the LORSAL algorithm; the contextual index $\alpha$; and
the rejection threshold $\rho$.  The regularization parameter
$\lambda$ describes the generalization capability of the classifier.
The contextual index $\alpha$ describes the contextual information;
$\alpha = 0$ means no contextual information and $\alpha = 1$ means no
classification information is taken in account.  The rejection
threshold $\rho$ denotes our confidence in the classification result;
lower values of $\rho$ denote low confidence in classification and
higher values of $\rho$ denote high confidence in classification.

\parcom{Training sets: different training sets to explore different
characteristics of the results}
To evaluate the parameters, we define two types of training sets,
based on the origin of the training samples: (1) A single image
training set composed of $k$ samples $S_k$, extracted from a test
image. This training set is used to train the classifier and is
applied to the entire image. (2) A training set $S_{k,k}$ containing
$k$ training samples from each image of a given set.  This training
set is used to evaluate the classifier in situations in which we have
no knowledge about the tissues.  Note that each of the $36$ H\&E
images not only contains a different set of tissues, but was also
potentially stained and acquired using different experimental
protocols, with no guarantee of normalization of the staining process.

\parcom{The rest of the parameters are empirically set}
The remaining parameters are set empirically \cred{according to the
  experts}.  The interscale
($v_\textrm{interscale}$) and intrascale ($v_\textrm{intrascale}$)
weights for the similarity graph construction are set to $4$ and $1$,
respectively, to achieve a ``vertical'' consistency in the multiscale
classification.
\cred{Larger values of the interscale when compared to the intrascale
  will enforce a higher multiscale effect on the segmentation: the
  different layers of the graph will be more similar to each other.}

The superclass misclassification cost $g$ is set to
$0.7$; the superclass consistency $\psi_c$ and rejection consistency
$\psi_r$ are set to $0.7$ and $0.5$, respectively, to ease transitions
into same superclass tissues and rejection, and to maintain a metric
interaction potential.
\cred{Larger values of the superclass consistency $\psi_c$ lead to smaller
  borders (in length) between elements of the same superclass, and
  smaller values lead to larger borders.
The value of the rejection consistency $\psi_r$ affects the length of the border of the rejected
areas (their perimeter): smaller values of $\psi_r$ lead to
disconnected rejected areas (with a large perimeter), usually thin rejection zones between two
different classes, whereas larger values of $\psi_r$ lead to connected
rejected areas (with a small perimeter), usually rejection blobs that
reject an entire area.
To achieve similar levels of rejected fraction, the rejection
threshold $\rho$ must accomodate the value of the $\psi_r$ as larger
values of $\psi_r$ mean more costly rejection areas.}


\paragraph{LORSAL Parameter Analysis}
\parcom{Expected behavior of the regularization parameter}
By varying the value of $\lambda$ in \eqref{eq:LORSAL}, we obtain
different regressors $\W_\lambda$ (one matrix of parameters per value
of $\lambda$.  We expect that by increasing the value of $\lambda$ up
to a point, a regressor with greater generalization capability can be
obtained, thus with increased classification performance.  However,
increasing $\lambda$ furthermore will lead to lower performance, as
the sparsity term in the optimization will overwhelm the data fit
term.  On the other hand, lower values of $\lambda$ will lead to an
overfitted regressor, that will cause loss of performance.


\parcom{Training set setup to evaluate the behavior of the parameters}
To evaluate the generalization capability of the classifier, we test
it with an entire data set training set $S_{75,75}$. 
With the entire data set training set created, each image is
classified by the following \textit{maximum a posteriori} classifier
for each of the regressors $\What_\lambda$ obtained for different
values of $\lambda$:
\begin{equation}
  \label{eq:class_no_context}
  \yhat_i = \arg\max_{\ell \in \mathcal{L}} p(y_i = \ell \mid \f_{i}, \W_\lambda)
\end{equation}
The overall accuracy is computed for each image, as well as the
sparsity of the regressor $\W_\lambda$.

\begin{figure}[htbp]
  \begin{center}
    \begin{tabular}{c}
      \includegraphics[width=.45\columnwidth]{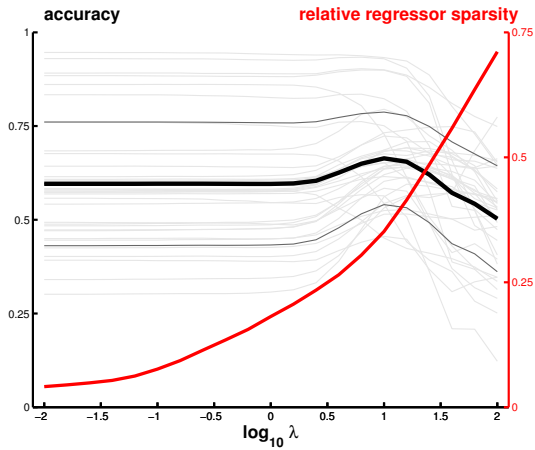}
    \end{tabular}
  \caption{
    \label{fig:LORSALlambda} LORSAL parameter analysis. Effect of
    $\lambda$ on the overall accuracy values and sparsity of $\W$.
\cred{Mean accuracy (in black), standard deviation (in gray), overlapped
with the results for all images.}
Note  the  three zones of accuracy behavior: no effect, increase,
    decrease. The maximum overall accuracy ($66.4 \%$) is obtained
    for $\lambda = 10$ with a value of relative regressor sparsity of
    $0.352$.}
  \end{center}
\end{figure}
\parcom{Observed behavior of the regularization parameters}
From Figure \ref{fig:LORSALlambda}, it is clear that there exist three
different zones of accuracy behavior with the increasing sparsity of
the regressor:
\begin{itemize}
\item For $0\leq \lambda \leq 1$ there is no effect --- the data term
  vastly outweights the regularization term;
\item For $1 \leq \lambda \leq 10$ there is an increase in
  classification performance --- increasing the regularization term
  will improve the generalization capability of the classifier;
\item For $\lambda > 10$ there is a decrease in classification
  performance --- increasing the regularization term will hamper the
  capability of the classifier.
\end{itemize}
We empirically choose $\lambda$ to be $10$, as it maximizes the overall
accuracy of the classifier.

\afterpage{\clearpage}
\paragraph{Effect of contextual index, and rejection threshold in
  the classification \cred{performance}}
\parcom{Observation problems. Classification quality as a viable solution.}
The inclusion of rejection in the classification leads to problems in
the measurement of the performance of the classifier.  As the accuracy
is measured only on the nonrejected samples, it is not a good index of
performance (the behavior of the classifier can be skewed to a very
large reject fraction that will lead to \cred{nonrejected} accuracies close to $1$).  To
cope with this, we use the quality of classification
$Q$~\cite{CondessaBK:15_rm}.
The intuition being that, by maximizing $Q$,
we maximize both the number of correctly classified samples not
rejected and the number of incorrectly classified samples rejected.
\begin{figure}[]
  \begin{center}
    \begin{tabular}{cc}
\includegraphics[width=.45\columnwidth]{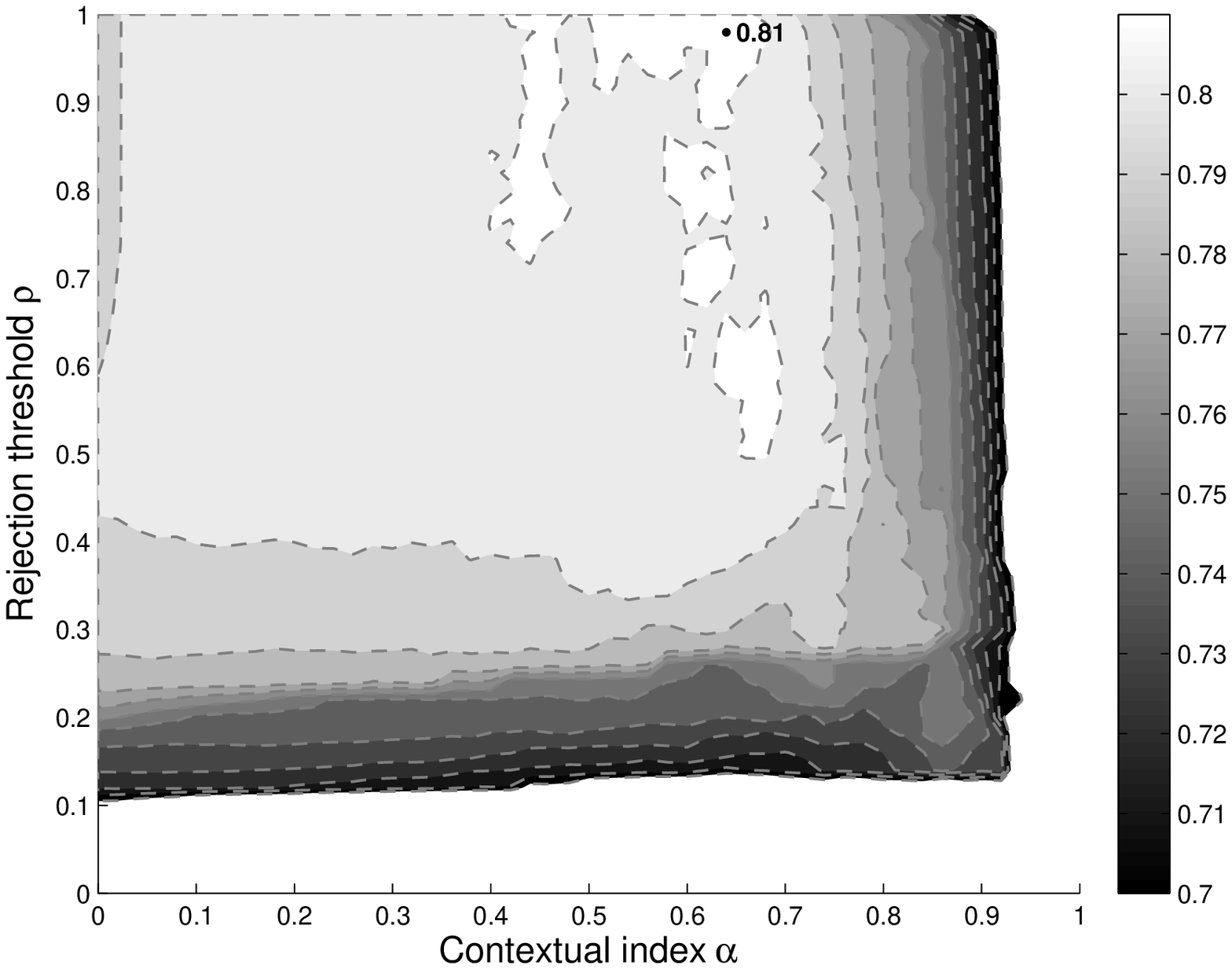} &
\includegraphics[width=.45\columnwidth]{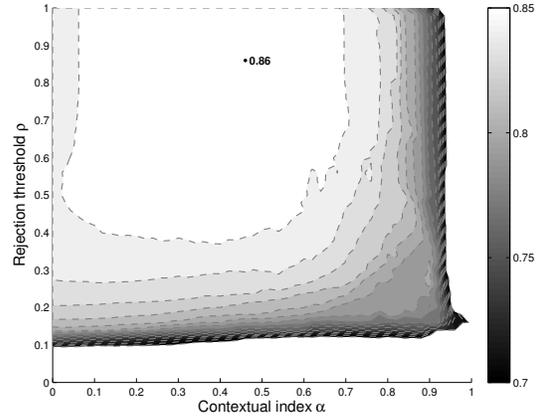} \\
\footnotesize{(a) $Q$ for $S_{60}$ (max. $Q$  $0.81$).}& 
\footnotesize{(b) $Q$ for $S_{120}$ (max. $Q$  $0.86$).} \\
\includegraphics[width=.45\columnwidth]{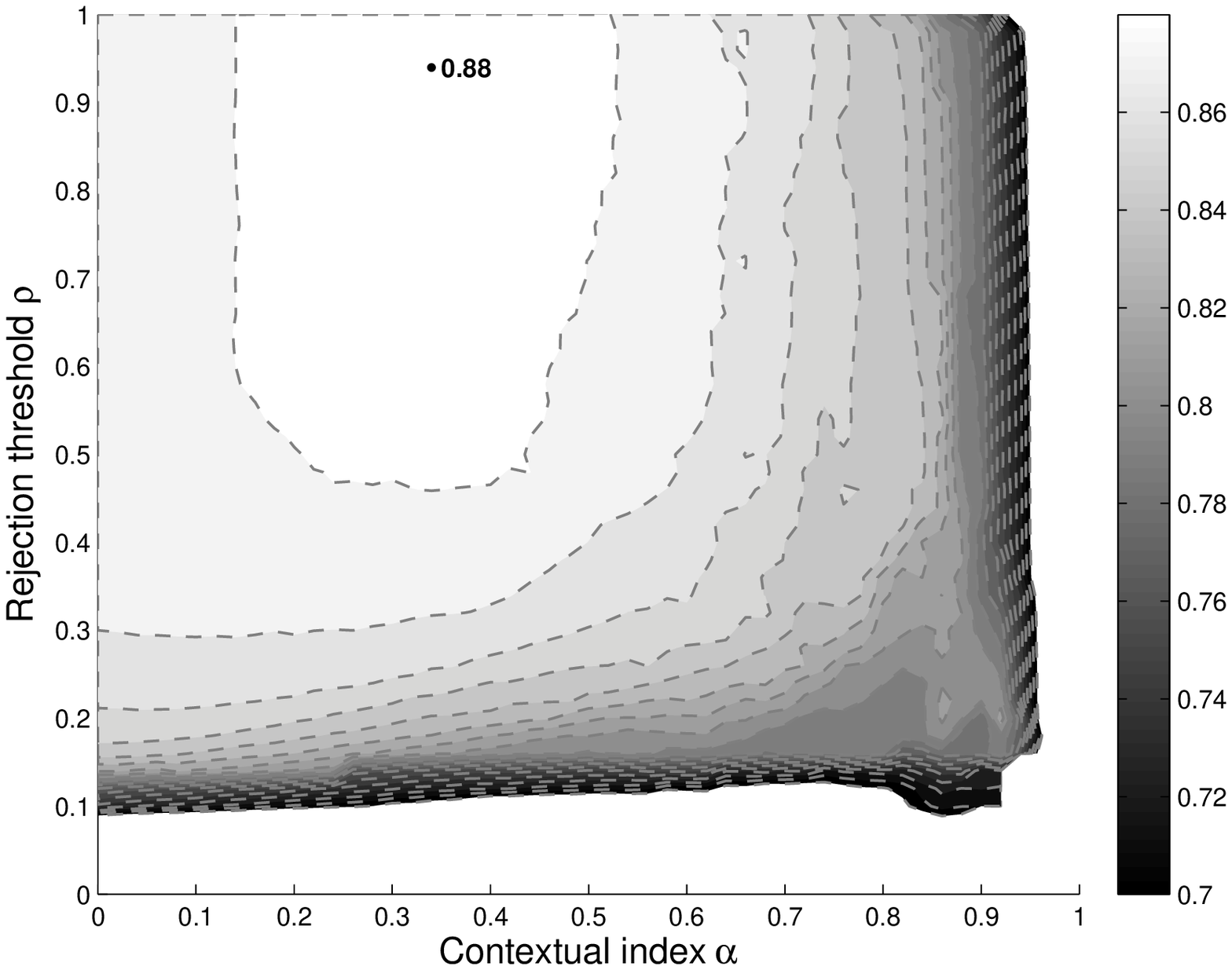} &
\includegraphics[width=.45\columnwidth]{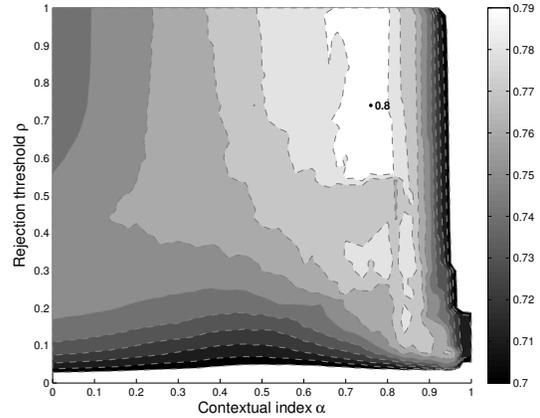}\\
\footnotesize{(c) $Q$ for $S_{240}$ (max. $Q$  $0.88$).} & 
\footnotesize{(d) $Q$ for $S_{60,60}$ (max. $Q$ $0.80$).} \\    
\end{tabular}
  \end{center}
  \caption{
    \label{fig:quality} Variation of quality of classification $Q$
    with the contextual index $\alpha$ and the rejection threshold
    $\rho$ for four different training sets. Adjacent contour lines
    correspond to a $0.01$ variation of $Q$.
\cred{It is clear a shift to lower dependency on rejection and contextual
information as the size of the training set, and consequently the
classifier performance, increases.}}
\end{figure}
\parcom{What we are observing with the different parameters}
By varying the value of the contextual index $\alpha$ in
\eqref{eq:problem_formulation}, we are weighting differently the role
of contextual information in the classification. For $\alpha = 0$, no
contextual information is used, equivalent to
\eqref{eq:class_no_context}, whereas for $\alpha = 1$, the problem
degenerates into assigning a single class to the entire image.  By
varying the value of the rejection threshold $\rho$ in
\eqref{eq:risk_definition}, we are assigning different levels of
confidence to the classifier, \textit{i.e.}, $\rho = 0$ is equivalent
to no confidence on the classifier (reject everything), whereas $\rho
= 1$ is equivalent to total confidence in the classifier (reject
nothing).

\parcom{The two parameters jointly interact}
As the contextual index $\alpha$ and the rejection threshold $\rho$
interact jointly, we now analyze the classification quality $Q$ for
different situations.

\parcom{Experimental setup description}
We test with three single image training sets $S_{60}$, $S_{120}$,
$S_{240}$, corresponding roughly to using $1.5\%$, $3\%$ and $6\%$ of
the samples of the image.  We test with an entire data set training
set $S_{60,60}$, on which only $3\%$ of the data set is composed of
samples from the test image.  For each type of training set, we use as
test images each of the $36$ images of the data set, presenting the
mean value of $Q$.

\parcom{Observed behavior}
From Figure \ref{fig:quality}, we can observe the variation of the
performance of the classifier with $\alpha$ and $\rho$ for different
situations.  The change from (a) to (c) corresponds to an increase in
the dimension of the training set.  Both the improvement of the
maximum value and the shift to lower values of contextual index and
higher values of rejection threshold can be explained by increasing
performance of the classification.  This means that a more reliable
classification is available, decreasing the need to use contextual
information and rejection.  On the other hand, (d) corresponds to an
extreme situation in which the training set is highly noisy, with only
$3\%$ of samples belonging to the test image. The high dependency of
contextual information in this case is clear.  The maximum value of
$Q$ is attained at lower values of the rejection threshold and higher
values of the contextual index.

\subsubsection{Parameter Selection}
\parcom{Composite rule for parameter selection}
As seen in Figure \ref{fig:quality}, the quality of classification
varies with the type of applications; applications for which the
training set is easier will lead to lower reliance on contextual
information and rejection, and harder training sets will lead to the
opposite.  In order to select a single set of parameters, we combine
the results of the four different training sets for each of the $36$
images, obtaining the average of the classification quality $Q$ and
nonrejected accuracy for the resulting $4 \times 36$ instances.  Our
motivation for the selection of the parameters is to maximize the
accuracy of the nonrejected fractions within a zone of high
classification quality.  To do so, we select the region of high values
of $Q$ ($Q$ higher than $99\%$ of its maximum value).  Then we select
the parameters that maximize the nonrejected accuracy, as seen in
Figure \ref{fig:par_sel}.
\begin{figure}[]
  \begin{center}
    \begin{tabular}{c}
\includegraphics[width=.5\columnwidth]{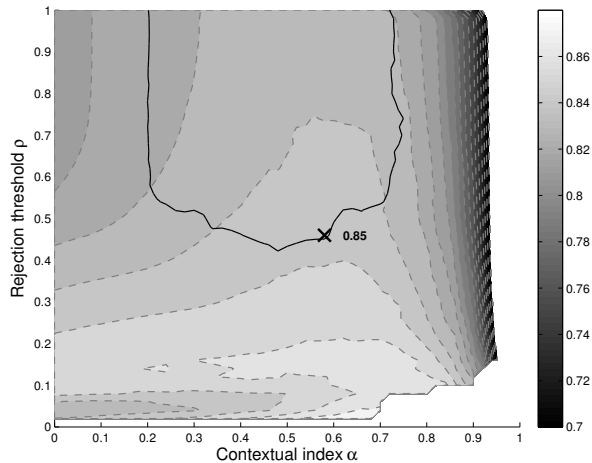} \\
\end{tabular}
  \end{center}
  \caption{ \label{fig:par_sel} Variation of nonrejected accuracy with
    the contextual index $\alpha$ and rejection threshold $\rho$. The
    dark line corresponds to the level set of quality of
    classification $Q$ equal to $99\%$ of its maximum value. The
    maximum nonrejected accuracy is $85\%$, corresponding to $\rho =
    0.46$ and $\alpha = 0.58$. The corresponding rejection fraction
    $r$ is $4.6\%$.}
\end{figure}

\begin{table}[ht]
\footnotesize
\begin{center}
\caption{\label{tab:ov_res} Classification and rejection performance metrics for the example images in Figure \ref{fig:class_res}.
Classification with rejection and context (white background), classification with context without rejection (green background), classification without context with rejection (red background), and classification without context and without rejection (brown background).}
\begin{tabular}{rrrrrr}
Image& \multicolumn{1}{l}{Nonrejected accuracy} & \multicolumn{1}{l}{Rejected fraction}  & \multicolumn{1}{l}{Rejectio  quality} & \multicolumn{1}{l}{Classification quality}& \multicolumn{1}{l}{Accuracy with no rejection} \\
\midrule
\multicolumn{6}{c}{Classification with rejection and context}\\
\midrule
$1$ &  ${ 0.701}$ & $0.347$ & $3.37$ & $0.662$&  \cellcolor{g!20} ${0.600}$ \\
$2$ &  $0.891$ & $0.067$ & $10.11$ & $0.868$ &\cellcolor{g!20} $0.862$ \\
$3$ &  $0.967$ & $0.140$ & $9.97$ &$0.866$ &\cellcolor{g!20}  $0.937$ \\

\midrule
\multicolumn{6}{c}{Classification with rejection without context}\\
\midrule
$1$  &  \cellcolor{r!20}$0.702$  &  \cellcolor{r!20}$0.370$  &  \cellcolor{r!20}$3.90$  &  \cellcolor{r!20}$0.673$  &  \cellcolor{s!20}$0.582$ \\
$2$ &  \cellcolor{r!20}$0.878$ & \cellcolor{r!20}$0.031$ & \cellcolor{r!20}$9.69$ &\cellcolor{r!20}$0.868$ & \cellcolor{s!20}$0.863$\\
$3$ &  \cellcolor{r!20}$0.936$  &  \cellcolor{r!20}$0.000$  &  \cellcolor{r!20}$3920$ &  \cellcolor{r!20}$0.936$ &  \cellcolor{s!20}$0.935$  \\

\bottomrule
\end{tabular}
\end{center}
\end{table}

\begin{figure}[]
  \begin{center}
    \begin{tabular}{ccc}
      \includegraphics[width=.3\columnwidth]{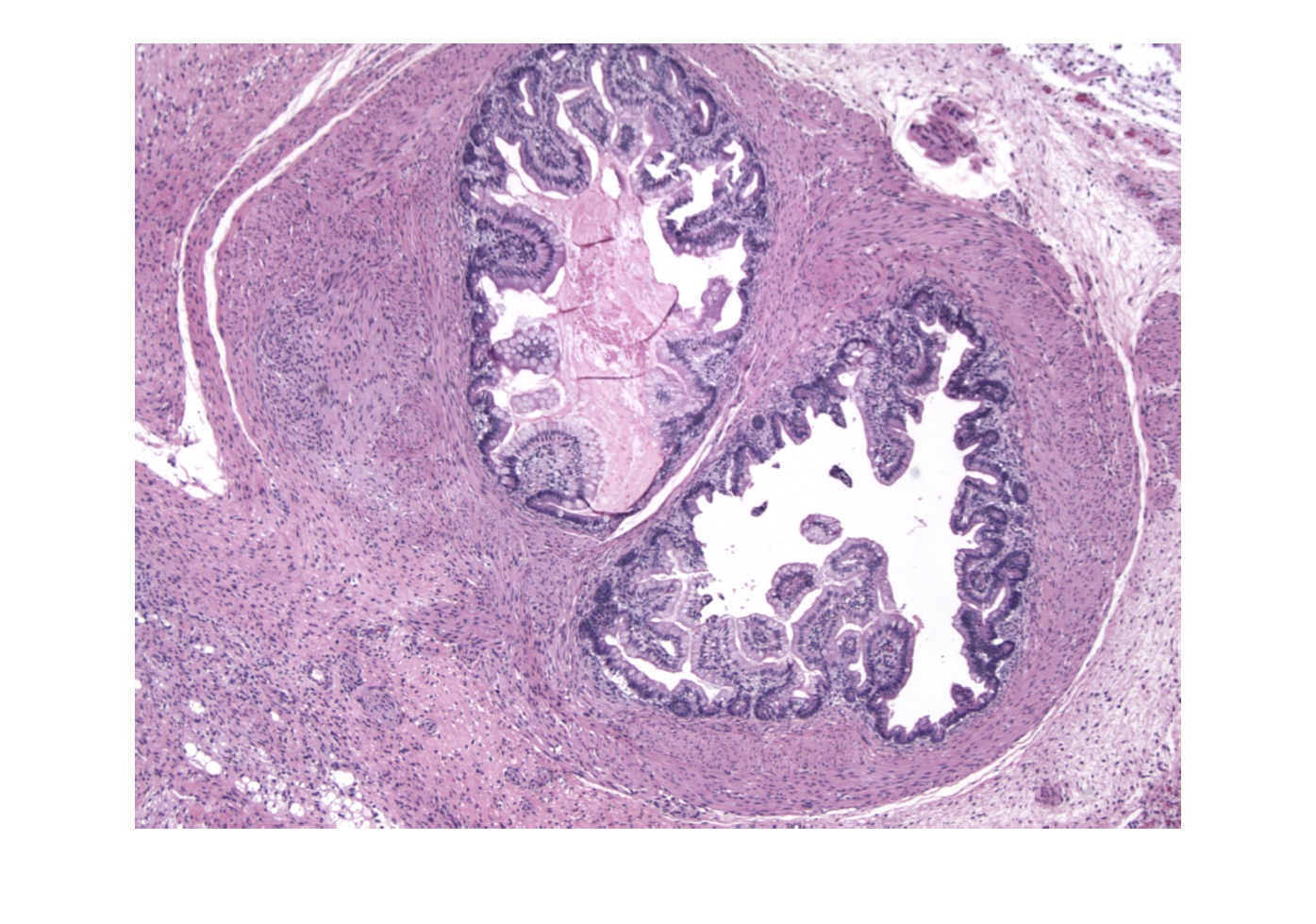} &
      \includegraphics[width=.3\columnwidth]{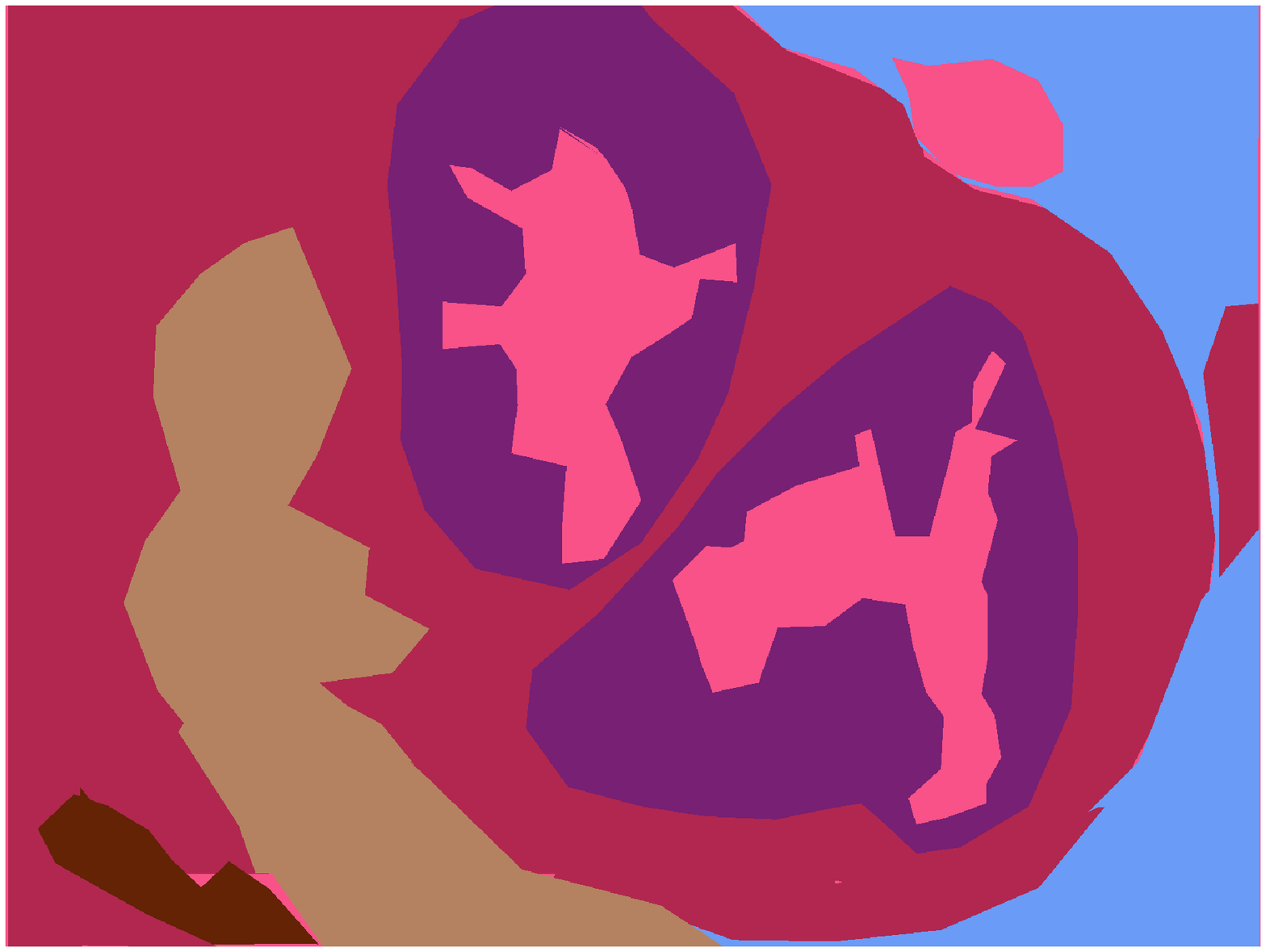} &
      \includegraphics[width=.3\columnwidth]{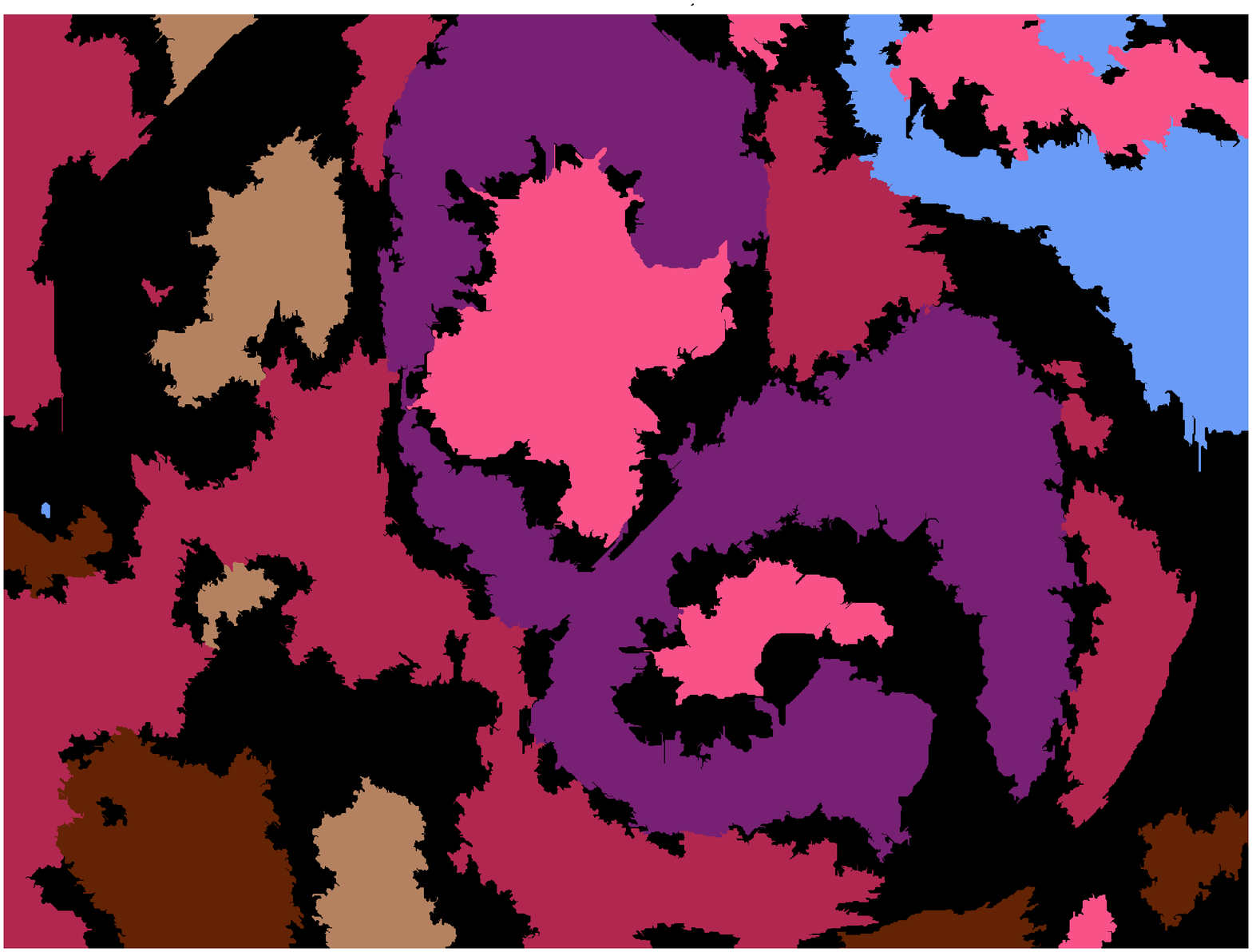}\\
      \includegraphics[width=.3\columnwidth]{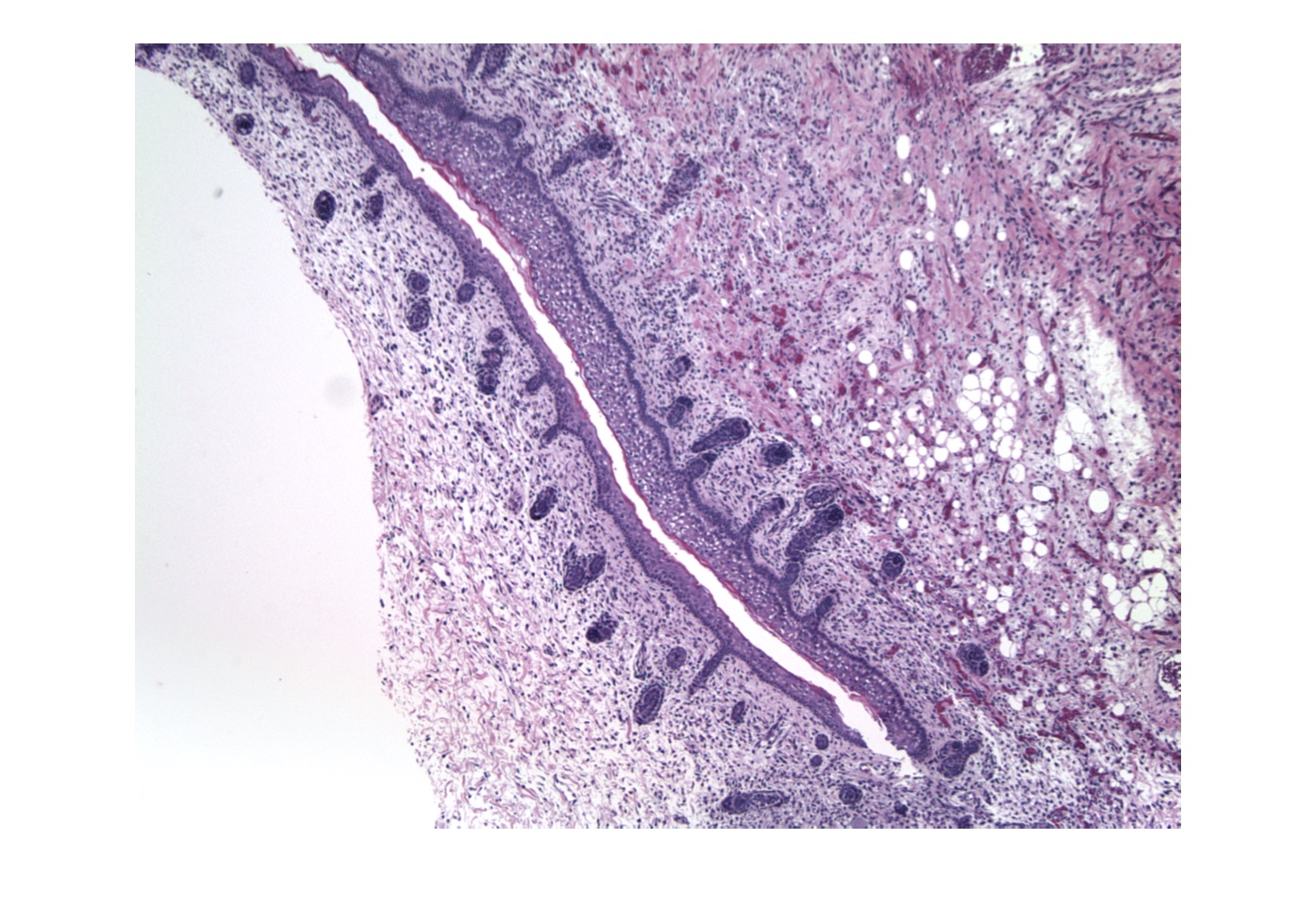} &
      \includegraphics[width=.3\columnwidth]{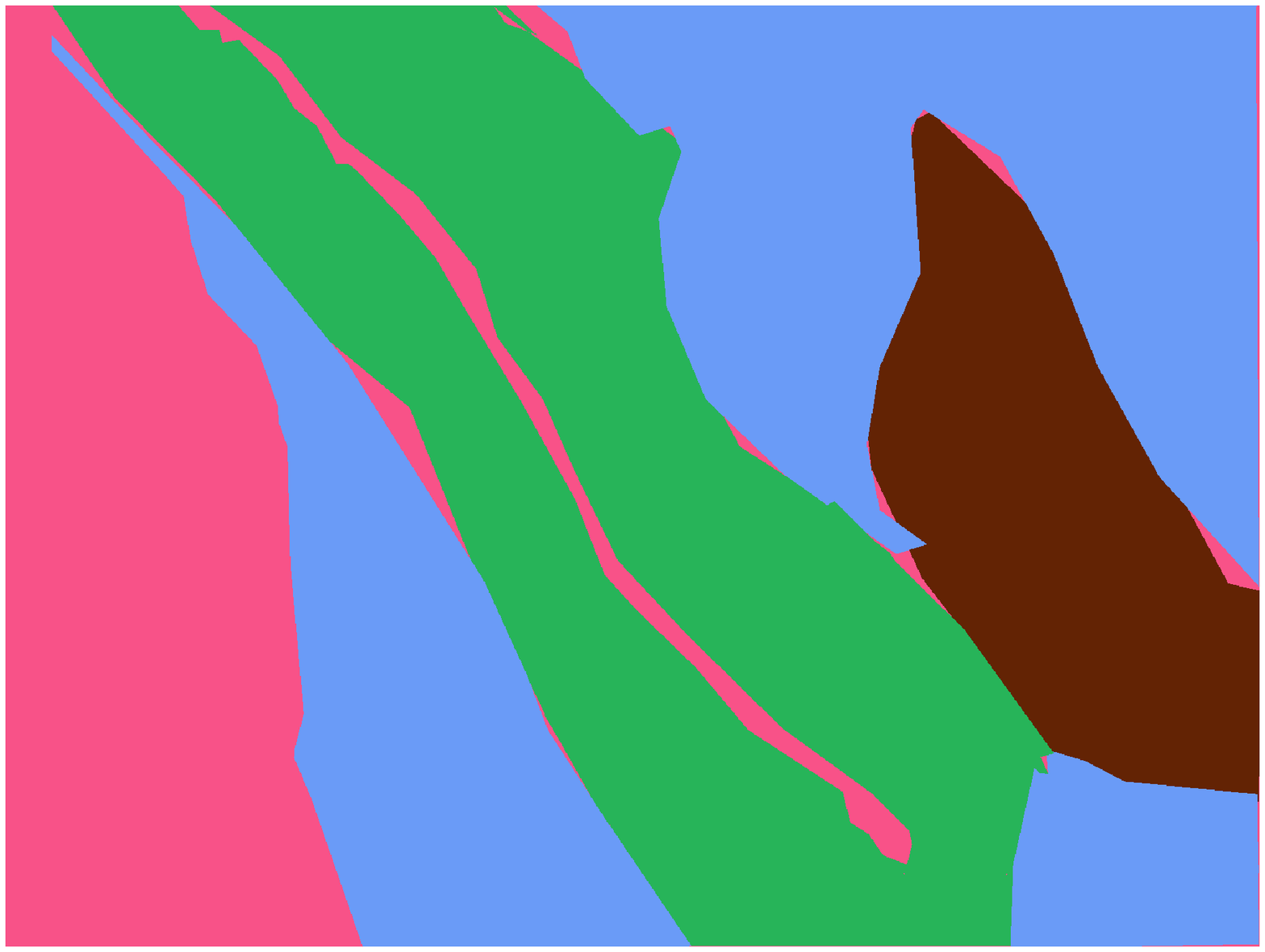} &
      \includegraphics[width=.3\columnwidth]{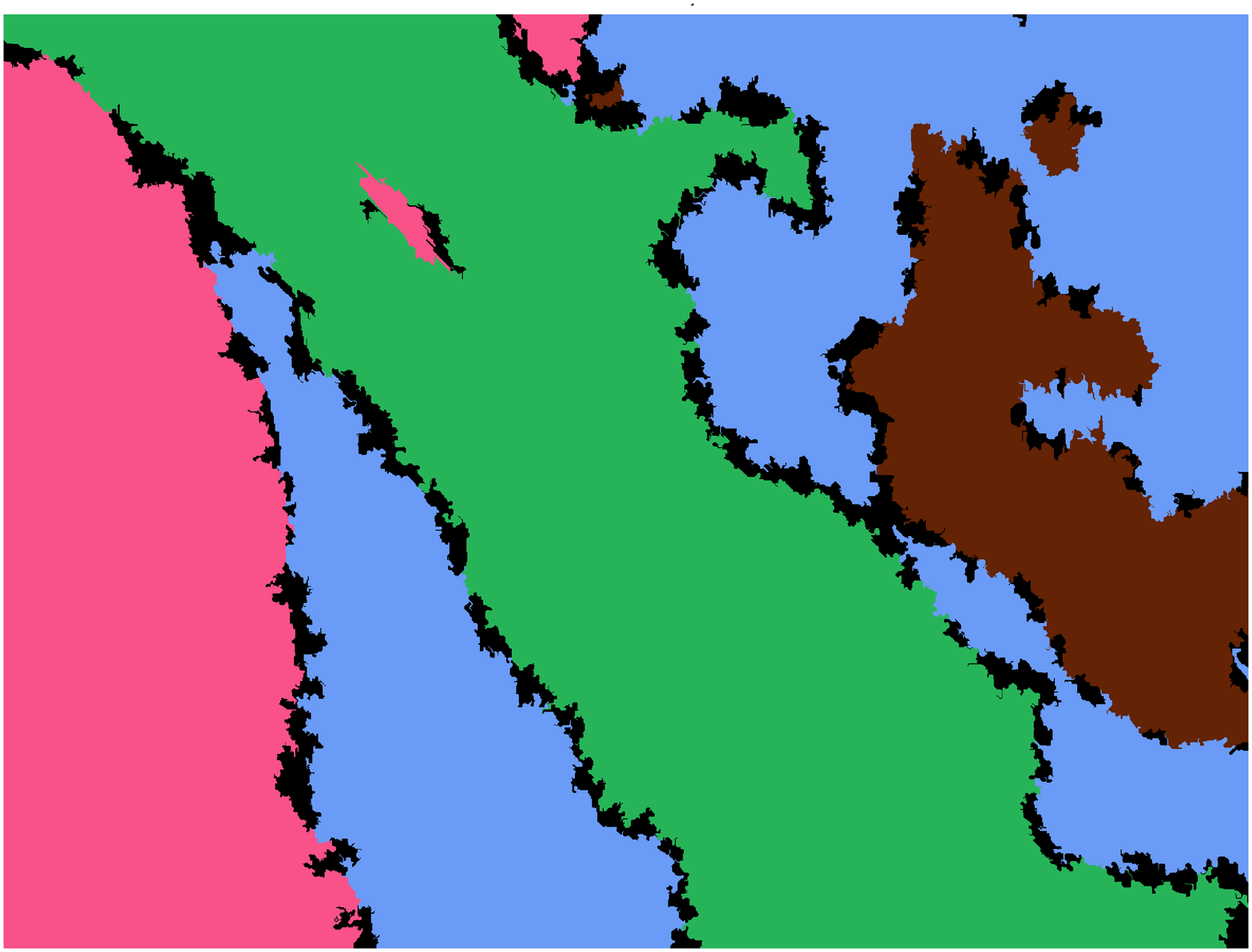}\\
      \includegraphics[width=.3\columnwidth]{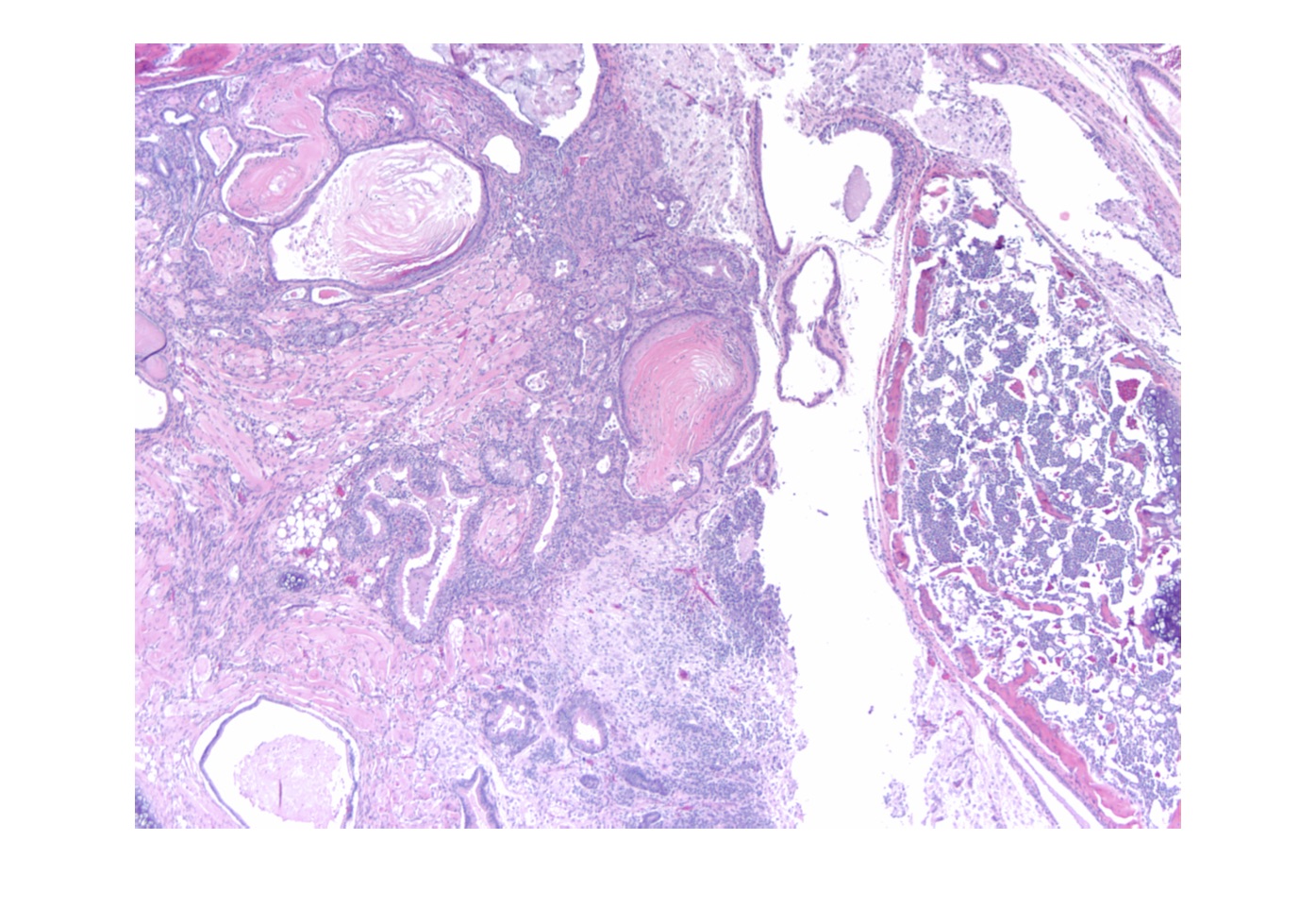} &
      \includegraphics[width=.3\columnwidth]{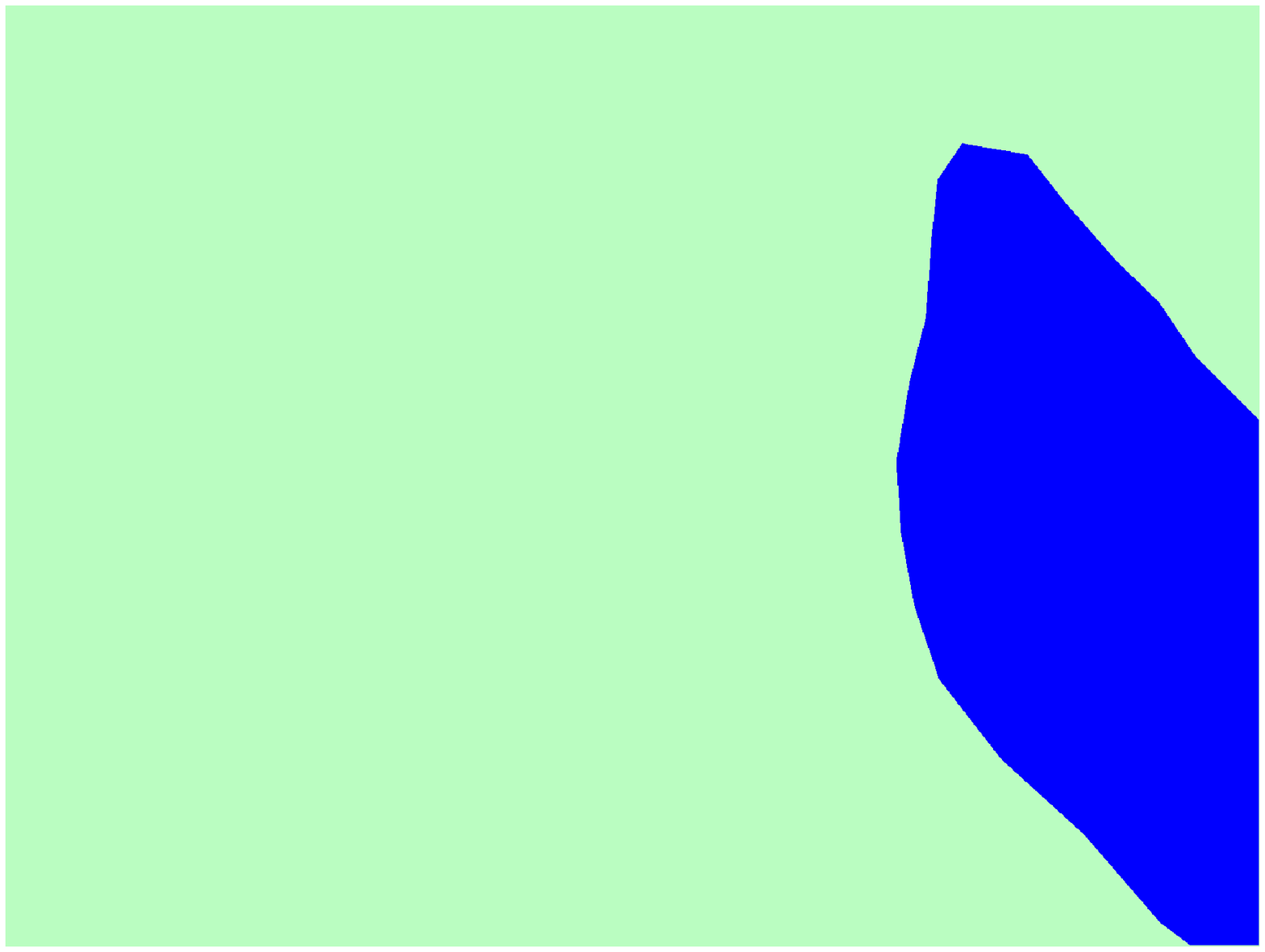} &
      \includegraphics[width=.3\columnwidth]{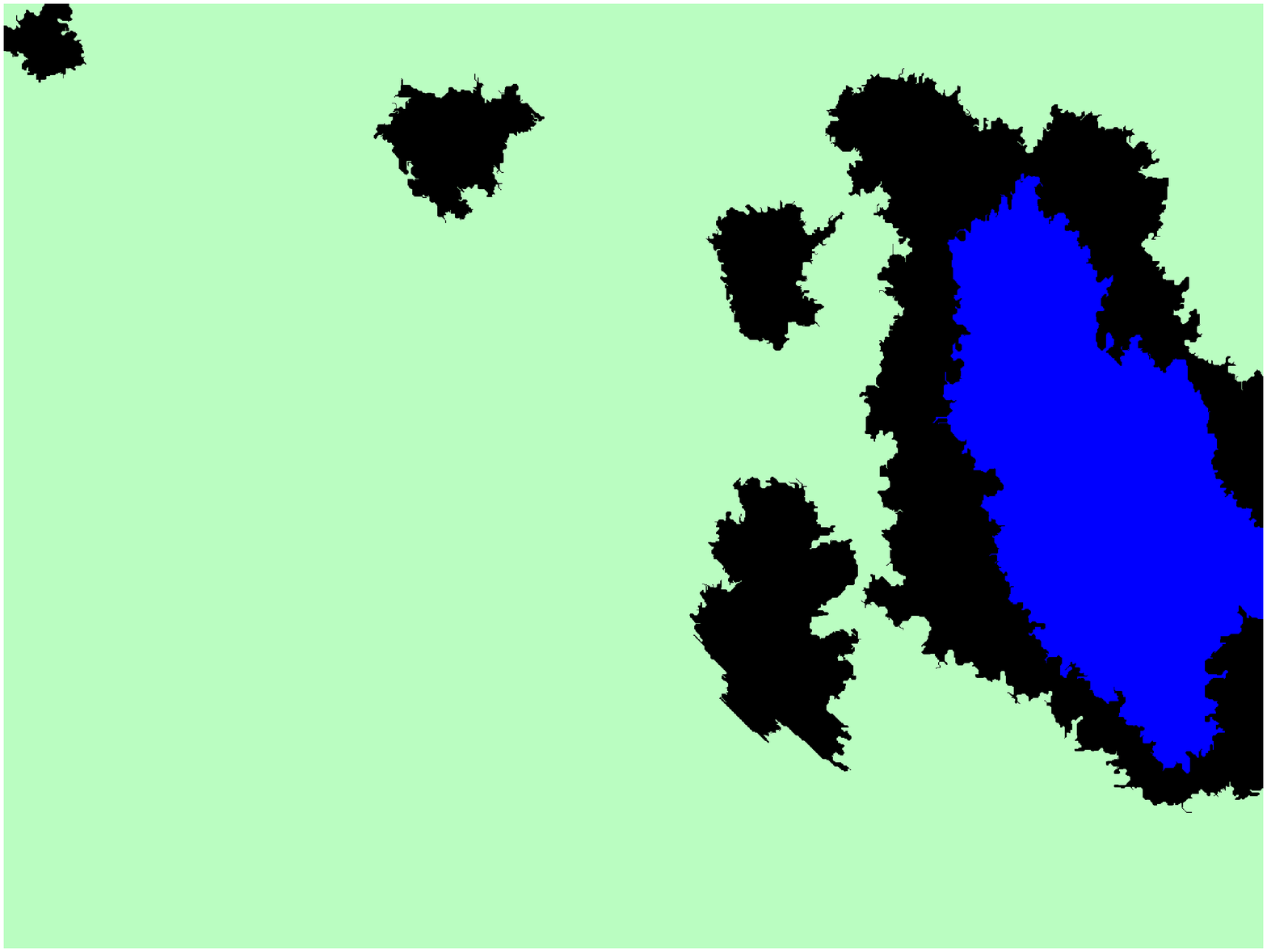}\\
      {\small (a) Original image.} &  {\small (b) Ground truth.} & {\small (c) Classification result.} 
\end{tabular}
  \caption{
    \label{fig:class_res} Example of classification results for H\&E
    stained samples of teratoma imaged at $40X$ containing multiple
    tissues: Image 1 (first row) background (light pink), smooth
    muscle (dark pink), gastro intestinal (purple), mature neuroglial
    (light brown), fat (dark brown); mesenchyme (light blue); Image
    2 (second row) background (light pink), fat (dark brown),
    mesenchyme (light blue), skin (green); Image 3 (third row)
    mesenchyme (light green); bone (dark blue). Rejected partitions
    are shown in black. The training set consists of: \cred{$5$ randomly chosen partitions per class (roughly $0.6\%$ of total) for image 1, } $120$ randomly
    chosen partitions (roughly $3\%$ of total) \cred{for image 2}, \cred{$20$ randomly chosen partitions (roughly $ 0.5\%$ of total) for image 3}; with the $\lambda$
    parameter set to $5$. 
}
  \end{center}
\end{figure}


\begin{table}[htbp]
\footnotesize
\begin{center}
\caption{\label{tab:class_res} Class-specific results for the example images in Figure \ref{fig:class_res}.}
\begin{tabular}{lrrrrrrrr}
\multicolumn{1}{c}{Tissue } & \multicolumn{1}{c}{Train }& \multicolumn{1}{c}{Test} & \multicolumn{1}{c}{Rejected }& \multicolumn{1}{c}{Rejection } & \multicolumn{1}{c}{Nonrejected}& \multicolumn{1}{c}{Classification}  &\multicolumn{1}{c}{Accuracy} \\
\multicolumn{1}{c}{type} & \multicolumn{1}{c}{samples}& \multicolumn{1}{c}{samples} & \multicolumn{1}{c}{samples}& \multicolumn{1}{c}{quality} & \multicolumn{1}{c}{accuracy}& \multicolumn{1}{c}{quality}  &\multicolumn{1}{c}{no rejection}  \\
\toprule
\multicolumn{8}{c}{Image 1} \\ 
\midrule
Other & $5$& $410$& $134$& $0.77$& $0.72$& $0.55$ &\cred{ $0.75$}\\ 
Fat & $5$& $54$& $1$& $0.00$& $0.94$& $0.93$ &\cred{$0.93$}\\ 
Gastrointestinal & $5$& $1036$& $170$& $3.90$& $0.91$& $0.83$ &\cred{$0.86$} \\ 
Smooth muscle & $5$& $1283$& $529$& $1.81$& $0.69$& $0.64$ &\cred{$0.58$} \\ 
Mesenchyme  & $5$& $454$& $174$& $4.04$& $0.53$& $0.66$ &\cred{$0.38$} \\ 
Mat. neuroglial & $5$& $369$& $143$& $1.82$& $0.35$& $0.53$ &\cred{$0.29$} \\ 
\midrule
\multicolumn{8}{c}{Image 2} \\ 
\midrule
Other & $30$& $885$& $24$& $5.80$& $0.91$& $0.90$ &\cred{$0.90$} \\ 
Fat & $13$& $510$& $48$& $4.54$& $0.77$& $0.75$ &\cred{$0.74$} \\ 
Skin & $36$& $1157$& $37$& $20.35$& $0.98$& $0.96$ &\cred{$0.97$} \\ 
Mesenchyme & $41$& $1268$& $127$& $6.17$& $0.86$& $0.83$ &\cred{$0.82$} \\
\midrule
\multicolumn{8}{c}{Image 3} \\ 
\midrule
Bone & $2$& $725$& $246$& $1.60$& $0.75$& $0.64$  &\cred{$ 0.69$}\\ 
Mesenchyme& $18$& $3195$& $319$& $11.27$& $1.00$& $0.91$  &\cred{$ 0.99$}\\ 
\bottomrule
\end{tabular}
\end{center}
\end{table}

\subsubsection{Results}
We present results of our method on a set of $3$ images from the data
set containing a different number of classes (as seen in Figure
\ref{fig:class_res}).  
\cred{The classifications are obtained with different training sets to
  illustrate different challenges.
In image $1$, to create a small and nonrepresentative training set, the
training set is composed of $5$ randomly chosen partition elements per
class (roughly $0.6\%$ of total).
In image $2$, to create a representative training set, the training
set is composed of $120$ randomly  chosen partition elements from the
entire image (roughly $3\%$ of total).
In image $3$, to create a small representative training set with high
class overlap, the training set is composed of $20$ randomly chosen
partition elements from the entire image (roughly $0.5\%$ of total).
In all cases, the $\lambda$ parameter is set to $5$, with the rest of
the parameters unchanged.}

We analyze both \cred{overall results (in Table \ref{tab:ov_res}) and class-specific results (in Table
\ref{tab:class_res})}.
The computation of the rejection quality is based on the results of
classification with contextual information and no rejection
(\emph{i.e.}  comparing the labeling with rejection to the labeling
resulting from setting the reject threshold $\rho$ to \cred{$1$} in
\eqref{eq:problem_formulation}).

\cred{In Table \ref{tab:ov_res}, we compare the performance of
classification with contextual information and rejection with context
only (obtained by setting $\rho = 1$) and with classification with
rejection only with optimal rejected fraction (obtained by sorting the partition elements according
to maximum\emph{ a posterior} probability and selecting the rejected
fraction that maximizes the classification quality).

\cred{
Comparing the performance results of classification with rejection using contextual information (white background in Tab. \ref{tab:ov_res}) with the results of classification with context only (red background in Tab. \ref{tab:ov_res}), the improvement in classification accuracy at the expense of introducing rejection is clear.
For images $1$ and $2$, this can be achieved at levels of classification quality higher than accuracy of context only, meaning that we are rejecting misclassified samples at a proportion that increases the number of correct decisions made (the underlying concept of classification quality).
For image $3$, due to the high accuracy of context only (and of the classification with no context and no rejection, brown background in Tab. \ref{tab:ov_res}), the increase in accuracy is at the expense of rejecting a comparatively large proportion of correctly classified samples, leading to a smaller value of classification quality.

Comparing the performance results of classification with rejection using contextual information with the results of classification with rejection only with optimal rejected fraction (red background in Tab. \ref{tab:ov_res}) the results are comparable for images $1$ and $2$, meaning we can achieve a performance improvement similar to the achieved by rejection with optimal rejected fraction through the introduction of context.
For image $3$, due to the high accuracy of classification with no context and no rejection (brown background in Tab. \ref{tab:ov_res}), the optimal rejected fraction is $0$, meaning that the increased accuracy is at the expense of rejecting a comparatively large proportion of correctly classified samples.
}

Analyzing the classification in Fig. \ref{fig:class_res}, the effects of combining rejection with contextual information are clear.
We obtain significant improvements for image $1$ by combining
classification with context with classification with rejection in
terms of classification quality and nonrejected accuracy, thus
revealing the potential of combining classification with rejection
with classification with context.
For image $2$, only the class boundaries are rejected, leading to high
values of overall rejection quality and class-specific rejection
quality.
In image $3$, it is clear the effect of noisy training sets (due to
the image characteristics), where a significant amount of the class
boundaries are rejected, and the classification quality is lower than
the accuracy of the original classification with no context and no rejection.}


\parcom{Usefulness of classification quality to evaluate the results}
Finally, we point to the usefulness of the classification quality $Q$.
By analysis of the classification quality, it is possible to compare
the performance of the classifier with rejection in different
situations and note how the performance will decrease as the
complexity of the problem increases (by increasing the number of
classes).

\section{Conclusions}
\label{sec:conclusion}
\parcom{Concluding paragraph}
We proposed a classifier where by combining classification with rejection with classification using contextual information we are able to 
increase classification accuracy.
Furthermore, we are able to impose spatial constraints on the rejection itself departing from the current standard of image classification with rejection. 
These encouraging results point towards potential application of this method in large-scale automated tissue identification systems of histological slices as well as other
classification tasks.

\section*{Acknowledgment}
The authors gratefully acknowledge support from the NSF through award
1017278 and the CMU CIT Infrastructure Award. This work was partially
supported by grant SFRH/BD/51632/2011, from   Funda\c{c}\~{a}o para a Ci\^{e}ncia e Tecnologia and the
      CMU-Portugal (ICTI) program, and by the project
      PTDC/EEI-PRO/1470/2012, from Funda\c{c}\~ao para a Ci\^encia e
      Tecnologia. Part of this work was presented in
      \cite{CondessaBCOK:13}.
We follow the principles of
reproducible research. To that end, we created a reproducible
research page available to readers \cite{CondessaBK:15_ICRCI}.

\ifCLASSOPTIONcaptionsoff
  \newpage
\fi



\end{document}